\UseRawInputEncoding
\documentclass{article}

\PassOptionsToPackage{numbers}{natbib}

\usepackage[preprint]{style}

\usepackage[utf8]{inputenc} 
\usepackage[T1]{fontenc}    
\usepackage{hyperref}       
\usepackage{url}            
\usepackage{booktabs}       
\usepackage{amsfonts}       
\usepackage{nicefrac}       
\usepackage{microtype}      
\usepackage{xcolor}         
\usepackage{amsmath} 
\usepackage{graphicx}
\usepackage{array}
\usepackage{multirow}
\usepackage{listings}
\usepackage{float}  
\usepackage{graphicx}
\usepackage{subcaption} 
\usepackage[table]{xcolor}
\usepackage{nicematrix} 
\usepackage{tcolorbox}
\tcbuselibrary{breakable, listings, skins}

\newtcblisting{rawpromptbox}[1][]{
    enhanced,
    breakable,
    colback=gray!5!white,
    colframe=gray!50!black,
    fonttitle=\bfseries,
    title=#1,
    boxrule=0.8pt,
    arc=3pt,
    listing only,               
    listing options={
        basicstyle=\ttfamily\small, 
        breaklines=true,            
        columns=fullflexible,
        keepspaces=true
    }
}
\title{EvoIR-Agent: Self-Evolving Image Restoration Agentic System via Experience-Driven Learning}
\author{%
  Kailin Zhuang$^{1}$, \quad Jiawei Wu$^{1}$, \quad Zhi Jin$^{1,2,3}$\thanks{Corresponding author.} \\
  $^{1}$School of Intelligent Systems Engineering, Shenzhen Campus of Sun Yat-sen University, China \\
  \texttt{zhuangklin@mail.sysu.edu.cn}
  \thanks{$^{2}$Guangdong Provincial Key Laboratory of Fire Science and Intelligent Emergency Technology, Shenzhen, China.}%
  \thanks{$^{3}$Guangdong Provincial Key Laboratory of Robotics and Digital Intelligent Manufacturing Technology, Guangzhou, China.}%
}

\begin{document}
\maketitle
\vspace{-15pt}
\begin{abstract}
Multimodal Large Language Model (MLLM)-driven image restoration agent demonstrates effectiveness in degradation coupling scenarios by flexibly selecting tools and determining removal orders. However, its zero-shot planning often fails without experience, necessitating severe trial-and-error overhead to achieve satisfactory outcomes. Currently, two paradigms are employed to address this issue, yet a dilemma persists: Training-based methods embed intrinsic experience into parameters, achieving high inference efficiency but lacking compatibility with new tools or degradations. In contrast, training-free methods utilize explicit experience storage for compatibility but still incur trial-and-error overhead due to naive experience. To resolve the dilemma, we propose \textbf{EvoIR-Agent}, which first systematically formulates the experience components of a training-free image restoration agent. Subsequently, a hierarchical experience pool is constructed, which enables coarse-to-fine guidance for diverse tools and removal orders. Furthermore, a self-evolving mechanism is introduced to update the pool from scratch using accumulated records, thereby greatly improving performance and efficiency. Extensive experiments reveal that \textbf{EvoIR-Agent} achieves a significant lead in the full reference metrics and yields a Pareto-optimal balance, which outperforms state-of-the-art methods in performance and is 61\% more efficient. Our codes are available at \url{https://github.com/rightleft-123/EvoIRAgent}.
\end{abstract}

\section{Introduction}
Image restoration (IR) is a low-level vision task that aims to recover a high-quality (HQ) image from a degraded low-quality (LQ) version. In recent decades, considerable progress has been made in addressing IR problems related to specific tasks \citep{zhang2017beyond, wang2018esrgan, tao2018scale, jin2025mb, wang2023fourllie}. However, in real-world scenarios, multiple degradations are often coupled, making restoration more challenging.

A promising direction for addressing this challenge is all-in-one image restoration (AiOIR), which aims to handle multiple degradations with a unified model \citep{jiang2025survey}. Most methods \citep{li2020all, valanarasu2022transweather, conde2024instructir, jiang2024autodir, lin2025jarvisir, wu2026gradient} address the degradation coupling problem by learning more general degradation feature representations. While significant strides have been made, they remain limited to a few degradations and are difficult to extend to new degradations without retraining.

Recent advances in MLLMs \citep{wang2024qwen2, zhang2024vision} have inspired a new direction in IR, namely Image Restoration Agent (IRA). The key is to leverage a MLLM-driven agent to select restoration tools and determine removal orders. However, general-purpose MLLMs encounter difficulties with zero-shot planning \citep{chen2024restoreagent, zhu2025intelligent} without experience-driven guidance, which results in severe trial-and-error overhead to reach satisfied restoration result. To resolve this issue, existing methods can be divided into two distinct paradigms. Training-based methods \citep{chen2024restoreagent, lu2025simplecall, zhang2026tir} internalize intrinsic experience into model parameters, offering high inference efficiency but lacking compatibility with new tools or degradations. Conversely, training-free methods \citep{zhu2025intelligent, jiang2025multi, NEURIPS2025_f0075fe4} utilize external containers for explicit experience storage, while highly compatible, they still result in considerable trial-and-error overhead owing to being guided by naive experience. Therefore, we aim to retain the compatibility of training-free methods while endowing the model with the inference efficiency of training-based methods. We identify that the considerable trial-and-error overhead in training-free IRAs stems from the underdeveloped experience mechanism. To systematically break this bottleneck, we revisit from two perspectives: (i) \textit{Do existing IRAs lack an experience mechanism altogether?} Not entirely. Recent frameworks \citep{zhu2025intelligent, jiang2025multi, NEURIPS2025_f0075fe4} already incorporate an experience mechanism. However, their experience components remain simplistic and static, preventing the agent from accepting comprehensive and dynamic guidance. (ii) \textit{Is it possible to transfer experience mechanisms from general-purpose agents to IRAs?} It is possible. Recent advancements \citep{zhao2024expel, zhang2025agentic, xu2025mem, fang2025lightmem} have demonstrated that explicit experience mechanisms drive self-evolving capability. However, it is unrealistic to directly apply it to IRAs without a domain-specific redesign.

\begin{figure}[t]
  \centering
  \includegraphics[width=\textwidth]{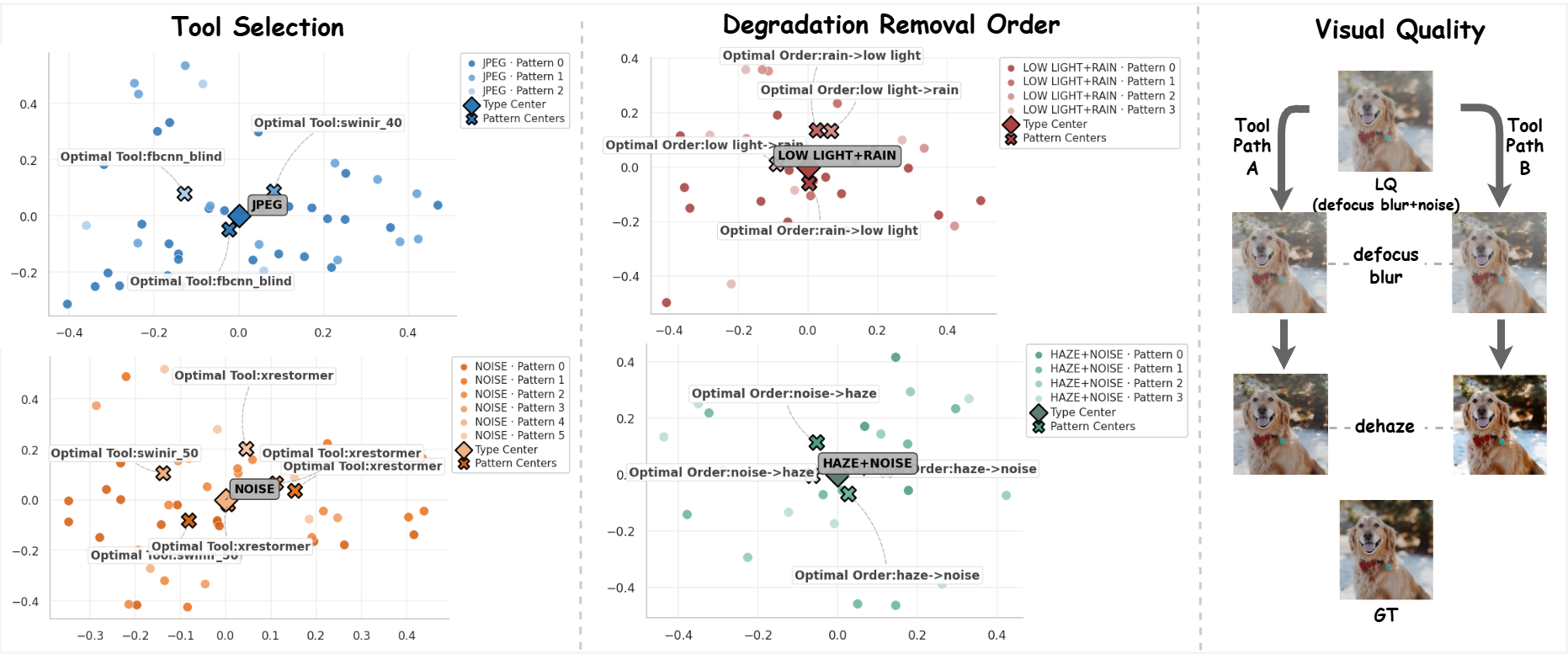}
  \caption{
  Illustration of the experience components in IRA. (Left\&Center) We extract embeddings of various degraded images using an image encoder \citep{huang2026llm2clip}. The scatter plots reveal that images sharing a degradation type may cluster into different degradation patterns. Crucially, the optimal \textit{Tool Selection} and \textit{Degradation Removal Order} are highly pattern-dependent. (Right) \textit{Visual Quality} demonstrates that divergent tool paths can achieve restoration in distinct visual quality.}
  \label{fig:introductions}
  \vspace{-15pt}
\end{figure}

Therefore, we introduce \textbf{EvoIR-Agent}, an image restoration agentic system explicitly designed to address the domain-specific needs. We first clearly identify the experience components in IRA as illustrated in Figure \ref{fig:introductions}. (1) \textit{Tool Selection}: The optimal tool depends not merely on broad degradation type, but on specific degradation pattern \citep{chen2024restoreagent}. Experience should be considered to map these patterns to their optimal tool; (2) \textit{Degradation Removal Order}: Similarly, in degradation coupling settings, the optimal order is also pattern-dependent; (3) \textit{Visual Quality}: Due to inherent attributes in different tools, multiple paths may restore the image, but result in different visual outcomes. 
Overall, current experience mechanism focuses on removal order, with a few addresses on tool selection and showing an absence in focusing on visual quality. Moreover, they concentrate on the broad degradation type, while neglecting to delve into the degradation pattern at a finer granularity. 

To systematically capture this coarse-to-fine information in conjunction with the aforementioned three experience components, we feature a hierarchical experience pool for storage. Inspired by \citep{zhangg, yu2025memweaver, pouransari2025pretraining}, this pool retrieves contextually relevant experience across multiple levels, providing the IRA with robust, prioritized experience guidance. Building upon the structural foundation, the pool is equipped with a self-evolving experience mechanism. Unlike prior static experiences, ours iteratively updates using newly accumulated records. This dynamic evolution enables the system from compatibility towards adaptability in tools, degradations, and data sources, thereby promoting performance and inference efficiency. Our main contributions are summarized as follows:

\textbf{(1) Systematic Formulation.} We reformulate the experience components in Image Restoration Agent by introducing \textit{visual quality} as a supplement. We also identify a limitation in current IRA experience, which solely focuses on degradation type while ignoring the degradation pattern.

\textbf{(2) Self-Evolving Architecture.} To effectively manage the formulated experience, we propose a hierarchical experience pool equipped with tailored retrieval methods. Building upon this structure, we introduce a self-evolving experience mechanism that updates the pool using accumulated records.

\textbf{(3) Empirical Evaluation.} Experiments demonstrate that our method outperforms state-of-the-art methods in performance and efficiency. Furthermore, the results validate the impact of the experience mechanism and its capacity for self-evolution, which endows the system with adaptability.

\section{Related Work}
\textbf{All-in-one Image Restoration.} All-in-one image restoration aims to handle multiple degradations within a unified model \citep{jiang2025survey}. AirNet \citep{li2020all}, PromptIR \citep{potlapalli2023promptir} and MiOIR \citep{kong2024towards} focus on designing a unified backbone to model diverse degradation distributions. DA-CLIP \citep{luo2023controlling} and InstructIR \citep{conde2024instructir} exploit degradation-aware vision-language priors for more flexible restoration control. Some methods leverage generative priors to enhance restoration quality. AutoDIR \citep{jiang2024autodir}, UniRestore \citep{chen2025unirestore}, and Defusion \citep{luo2025visual} formulate restoration in a diffusion framework. Later methods explore more general restoration representations \citep{wu2026gradient, tang2026learning}. However, these works are constrained to a limited number of degradations, and their capacity for extension is limited.

\textbf{Image Restoration Agent.} Existing image restoration agent approaches can be broadly divided into training-based and training-free methods. Training-based methods, such as RestoreAgent \citep{chen2024restoreagent} and Q-Agent \citep{zhou2025q}, equip the MLLM with degradation perception and planning capability through fine-tuning. Restore-R1 \citep{lu2025simplecall}, TIR-Agent \citep{zhang2026tir}, and PaAgent \citep{wang2026paagent} follow in leveraging reinforcement learning (RL) for efficiency planning. In contrast, training-free methods focus on system design. AgenticIR \citep{zhu2025intelligent}, MAIR \citep{jiang2025multi}, and 4KAgent \citep{NEURIPS2025_f0075fe4} place significant emphasis on agentic workflow orchestration. Nevertheless, these works neglect a robust experience mechanism.

\textbf{Experience Mechanism in General-Purpose Agent.} Recent methods have demonstrated that explicit experience can enhance the agent's capabilities \citep{zhang2025survey, huang2026rethinking}. Early agentic systems, such as ExpeL \citep{zhao2024expel}, implement this idea by collecting and reflecting on trajectories in natural language, then retrieving distilled information at test time. Subsequent works \citep{xiong2025memory, xia2025experience} examine not only how to store experience, but also how to manage it. Finally, FLEX \citep{cai2025flex} and SimpleMem \citep{liu2026simplemem} treat experience as the engine of continual adaptation. Overall, existing methods suggest that the experience mechanism in agents is evolving from simple episodic replay toward selective memory management, promising the system's ability to evolve and adapt.

\section{Methodology}
\label{sec:Methodology} 
\begin{figure}[H]
\centering
\begin{minipage}{0.5\columnwidth}
This section introduces the closed-loop framework of the \textbf{EvoIR-Agent}. As demonstrated in Figure \ref{fig:inference_framework}, we begin with the outer loop by establishing the five-process inference workflow (Sec.~\ref{subsec:experience_driven_inference_workflow}). Subsequently, we shift to the inner loop and begin with mathematically formulating the experience. By encapsulating three core components—tool selection, degradation removal order, and visual quality—we model the experience by a priority-driven mapping function $p$ (Sec.~\ref{subsec:experience_formulation}). To manage these components, we design a hierarchical experience pool structured across three granularities. This architecture practically instantiates $p$ specifically for each level, providing coarse-to-fine guidance during inference (Sec.~\ref{subsec:hierarchical_experience_pool}). Ultimately, we introduce a self-evolving experience mechanism that functions in a batch-wise manner. This mechanism enables the continuous updating of $p$ from accumulated records, thereby ensuring performance efficiency and enabling adaptability in the system. (Sec.~\ref{subsec:self-evolving_experience_mechanism}).
\end{minipage}
\hfill
\begin{minipage}{0.44\columnwidth}
  \includegraphics[width=\linewidth]{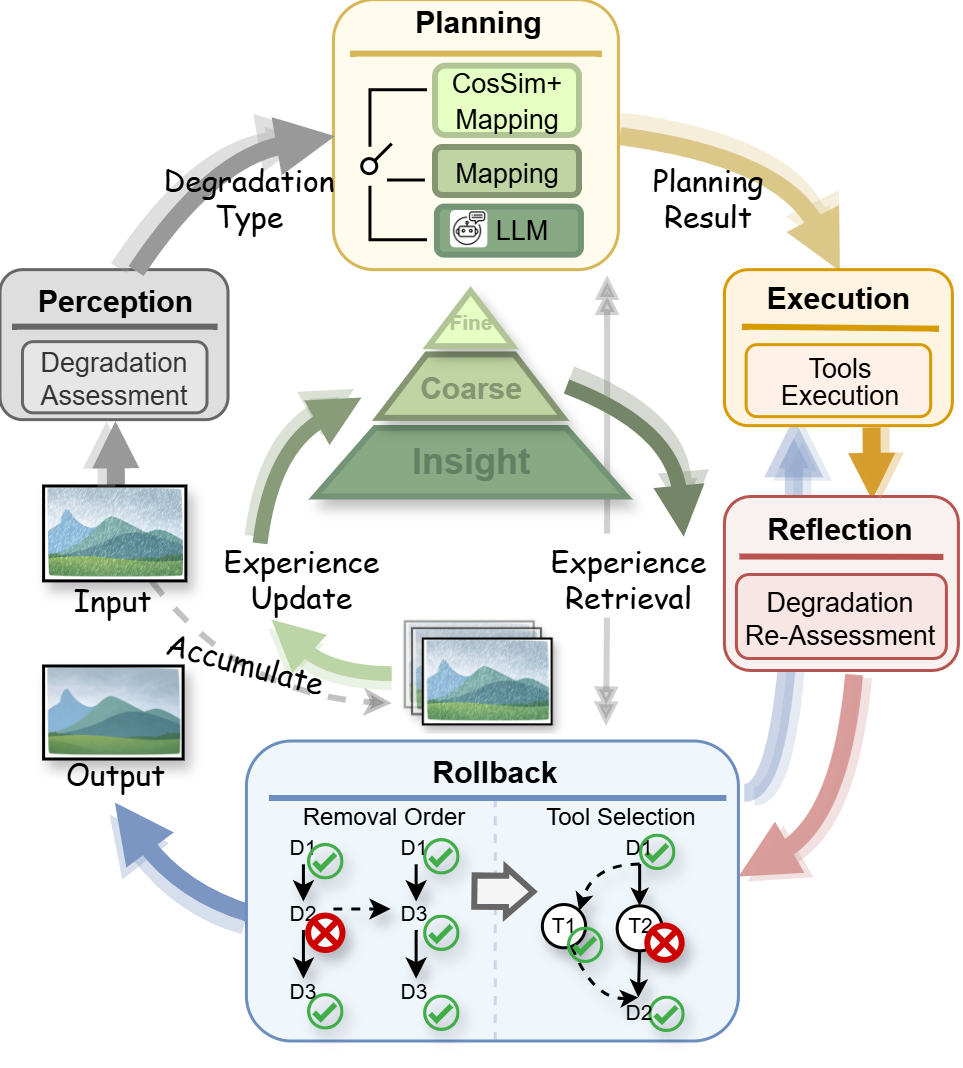}
  \caption{Overview of the experience-driven inference workflow in \textbf{EvoIR-Agent}.}
  \label{fig:inference_framework}
\end{minipage}
\end{figure}

\subsection{Experience-Driven Inference Workflow}
\label{subsec:experience_driven_inference_workflow}

The five-process inference workflow adopted in prior training-free IRAs \citep{zhu2025intelligent} is maintained. Formally, given a degraded image $I_D$, the \textit{Perception} process identifies the degradations, denoted as $D = \{d_1, d_2, \dots\}$. Based on the result, the \textit{Planning} process performs decision-making with the degradation removal order $O$, which is defined as a permutation over $D$, i.e., $O \in \Pi(D)$. Adhering to this plan, the \textit{Execution} process utilities a series of restoration tools for specific degradation $d_i$, defined as $T = \{\tau_{d_1}^{1}, \tau_{d_1}^{2}, \dots, \tau_{d_2}^{1}, \dots\}$ to obtain the restored image $\hat{I}_D$. The \textit{Reflection} process is then performed to evaluate if all $D$ have been eliminated on $\hat{I}_D$. If so, the process terminates and outputs the image. Otherwise, the \textit{Rollback} process is triggered. The utilization of the experience pool enables the MLLM to accept robust guidance during the \textit{Planning} and \textit{Rollback} processes, and facilitates the management of $O$ and $T$ usage. 

\subsection{Experience Formulation}
\label{subsec:experience_formulation}
We first define the aforementioned concept of visual quality. Observations from the restored images $\hat{I}_D$ from prior works (detailed in Appendix \ref{appendix:a1_experience_proving}) reveal a critical phenomenon: distinct tool selections can yield restored images that exhibit divergent visual quality. We conceptualize these differences as a \textit{Preference} requirement, denoted by $Q$. It signifies a subjective preference interpretation that can be specified by humans and supported by different image quality assessment (IQA) metrics. In this work, \textit{Fidelity} and \textit{Perception} are specified as \textit{Preference} requirements.

Subsequently, we mathematically formulate the guidance provided by the experience in the priority-driven mapping function $p$. The objective of \textit{Tool Selection} is:
\begin{equation}
    \tau^{*} = p_{tool}\big(T(d)\mid Q, d, H\big),
\end{equation}
where $T(d) \subseteq T$ indicates the candidate tool set corresponds to a degradation $d$, and $H$ contains the historical information during the decision-making process in the inference stage.

In cases of degradation coupling setting, the objective of determining the \textit{Degradation Removal Order}, denoted by $O$, should be established in conjunction with the relevant restoration tools $T(D)$:
\begin{equation}
    o^{*} = p_{order}\big(O \mid Q,  T(D), D, H\big).
\end{equation}
\textbf{Experience Foundation.} It should be noted that the number of permutations in $O$ increases exponentially as $D$ and $T$ rise, which makes direct experience modeling inefficient and redundant. Therefore, we propose to simplify the condition on experience and form it as a foundation. The details are provided in Appendix \ref{appendix:a1_experience_proving}. Here's a summary:
\begin{itemize}
    \item It is imperative to formulate $p_{tool}$ before $p_{order}$, since $p_{tool}$ can be considered as $p_{order}$ during $|D| = 1$. Subsequent sections will denote $p$ for unified representation.
    \item During the process of experience formulation, when specifying the preference $Q$, we decoupled $p_{tool}$ and $p_{order}$, which reduces the joint optimization from the $\prod_{d_i \in D} \mathcal{T}(d_i)$ search space down to a purely $D$-centric problem for the priority function $p_{order}$. 
    \item During rollback in inference time, $O$ is revised first with the optimal tool $\tau^{*}$ derived from $p_{tool}$, followed by the refinement of $T$ according to $p_{tool}$.
\end{itemize}

\subsection{Hierarchical Experience Pool}
\label{subsec:hierarchical_experience_pool}
In light of the aforementioned formulation and the findings presented in Figure \ref{fig:introductions}, the optimal $\tau^*$ and $o^*$ depend not only on the degradation type but also on the degradation pattern. Therefore, we draw inspiration from the hierarchical experience pool used in general-purpose agents, which effectively balances information across different granularities. Adopting this structure leads to a distinct instantiation of the priority-driven mapping function $p$:
\begin{figure}[t]
  \centering
  \includegraphics[width=\textwidth]{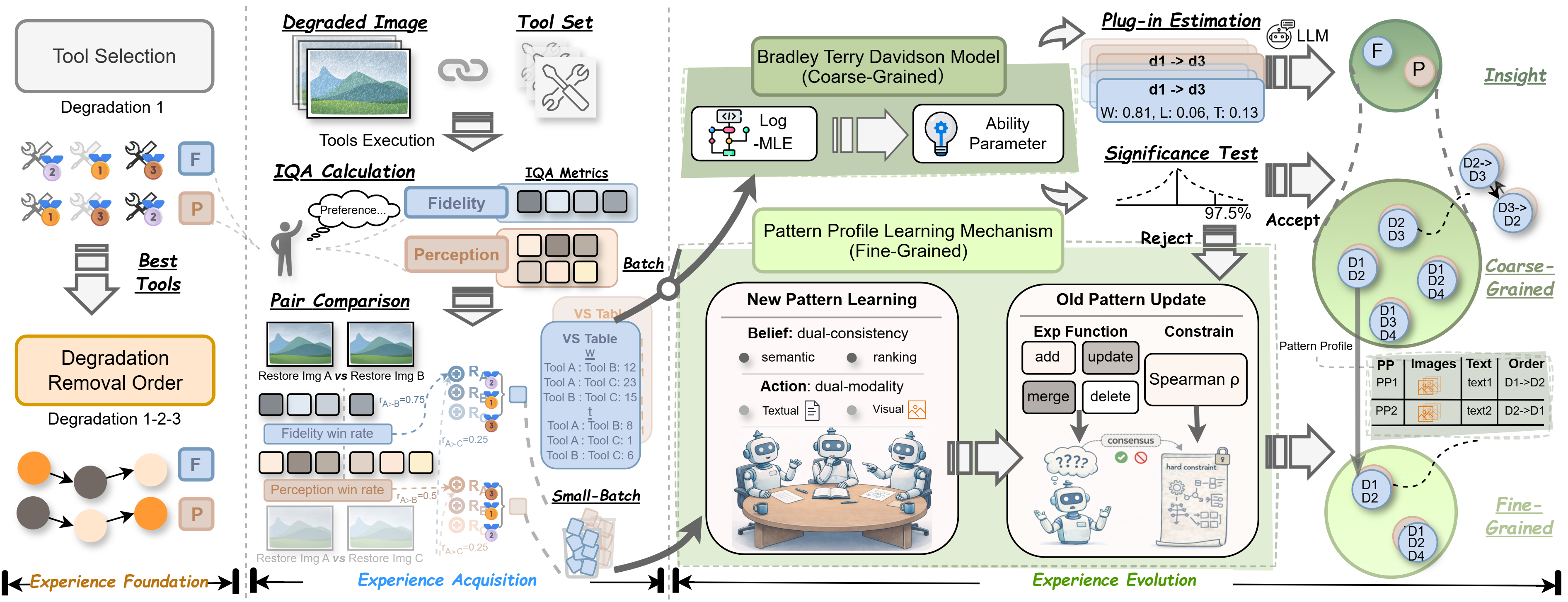}
  \caption{
  Overview of the inner closed-loop experience mechanism in \textbf{EvoIR-Agent}. (Left) Experience pool works as a priority-driven mapping problem. (Middle) Using priority-aware modeling to generate an atomic experience record. (Right) Batch-wise experience evolving mechanism for the hierarchical pool to update the priority function.
  }
  \label{fig:experience_architeture}
\end{figure}

\textbf{Insight Level (Preference-oriented):} At the highest level, $p$ is instantiated by a LLM. For given $Q$ and $D$, the LLM leverages distilled textual experience to guide the planning on $o$.

\textbf{Coarse-grained Level (Degradation Type-oriented):} This level instantiates $p$ into a direct key-value mapping without the MLLM involved. Utilizing $Q$ and $D$ as the query,  the value is considered as $o$.

\textbf{Fine-grained Level (Degradation Pattern-oriented):} According to Figure \ref{fig:introductions}, a degradation type may cluster into different degradation patterns. This level still instantiates $p$ into a key-value mapping, while combined with a retrieval strategy. The core intuition here is that a similar degradation pattern exhibits a similar priority function $p$. We first anchor these specific patterns using a \textit{pattern profile}:
\begin{equation}
    P = \left( \{ I_D^{(i)} \}_{i=1}^m,\; text \right), 
\end{equation}
where $\{ I_D^{(i)} \}$ is a supporting set of degraded images characterizing the pattern, and $\text{text}$ is its semantic description. (The construction of $P$ is detailed in Section \ref{subsec:self-evolving_experience_mechanism}). During inference, given an input image $I_D$ along with $Q$ and $D$, we execute a two-stage cascade retrieval. First, we employ a CLIP encoder $e(\cdot)$ \citep{huang2026llm2clip} to recall the Top-K relevant profiles based on cosine similarity:
\begin{equation}
    \operatorname{CosSim}(P, I_D)=
    \frac{e(P)\cdot e(I_D)}{\lvert e(P)\rvert \, \lvert e(I_D)\rvert}.
\end{equation}
Subsequently, the MLLM evaluates the Top-K candidates ${text}$ against the input image to identify the most accurate profile $P^*$. Ultimately, $P^*$ is utilized as a query condition, value is considered as $o$.

In light of the preceding experience foundation, the aforementioned three levels are characterized by the automatic selection of  $\tau$ on each $d$ with preference $Q$ according to the coarse-grained level.

\subsection{Self-Evolving Experience Mechanism}
\label{subsec:self-evolving_experience_mechanism}
Based on the previous formulation of each experience and the structure of the experience pool, this section explains how the entire self-evolving experience mechanism is applied to the experience pool and operates in generating the latest priority-driven mapping function $p$.

\textbf{Experience Acquisition.} As illustrated in the middle panel of Figure \ref{fig:experience_architeture}, we undertake an exhaustive exploration of all potential outcomes to establish the comprehensive priority-aware model to generate an atomic experience record, which bootstraps the subsequent evolution stage. 

Within an atomic experience record, a degraded image $I_D$ with degradation types $D$ and its corresponding tools $T(D)$ are employed, thereby yielding a set of restoration images:
\begin{equation}
 \hat{I}_D = \{ \hat{I}_{D}^1, \dots, \hat{I}_{D}^N \}, 
 \text{ where } N = 
 \begin{cases} 
    |T(D)|, & |D|=1 \\ |D|!, & |D|>1
\end{cases}
\label{search}
\end{equation}
Subsequently, the results are evaluated using IQA metrics according to preference $Q$. A pairwise comparison is performed on each metric in regard to the restoration images $\hat{I}_D$:
\begin{equation}
    r_{i \succ j} =
    \frac{1}{|M|}\sum_{m \in M} \mathbb{I}_m(i \succ j),
    \quad
    \mathbb{I}_m(i \succ j) =
    \begin{cases}
        1, & m \in M\uparrow \text{ and } m(\hat{I}_{D}^i) > m(\hat{I}_{D}^j)\\
        1, & m \in M\downarrow \text{ and } m(\hat{I}_{D}^i) < m(\hat{I}_{D}^j),\\
        0, & \text{otherwise}
    \end{cases}
\end{equation}
where $M\uparrow$ and $M\downarrow$ denote the higher-is-better and lower-is-better metrics, respectively. Based on these, we derive the win rate pair $r_{i \succ j}$ along with the win/loss/tie statistics $w_{ij}$, $l_{ij}$, and $t_{ij}$, where $r_{i \succ j}>0.5$ indicates a vote for $i$, $r_{i \succ j}<0.5$ indicates a vote for $j$, otherwise, considered a tie.

Following this, the average win rate of each tool is calculated, and its ranking is derived via sorting:
\begin{equation}
    R_i =
    \frac{1}{|\hat{I}_D| - 1}
    \sum_{j \neq i} r_{i \succ j},
    \qquad
    \operatorname{rank}_i = \operatorname{rank}_{\downarrow}(R_i).
\end{equation}
\textbf{Experience Evolution.} As illustrated in the right panel of Figure \ref{fig:experience_architeture}, once a set of $B$ atomic experience records has been collected for $D$, the process will be triggered on the batch. Ultimately, the designated level of  $D$ will be used in inference.
 
In the coarse-grained level,  the self-evolving process commences with the utilization of the latest win/loss/tie counts $w_{ij}$, $l_{ij}$, and $t_{ij}$ in $B$ rounds in conjunction with the preceding counts. Leveraging these statistics, the \textit{Bradley--Terry--Davidson (BTD)} model \citep{davidson1970extending} is adopted for the purpose of parameterizing the win/loss/tie relations in a probabilistic manner:
\begin{equation}
    P(i \succ j)=\frac{e^{\theta_i}}{e^{\theta_i}+e^{\theta_j}+2\nu e^{(\theta_i+\theta_j)/2}},\quad
    P(i = j)=\frac{2\nu e^{(\theta_i+\theta_j)/2}}{e^{\theta_i}+e^{\theta_j}+2\nu e^{(\theta_i+\theta_j)/2}},
\label{eq:btd_model}
\end{equation}
where $\nu \ge 0$ denotes the tie-intensity parameter, and $\theta$ defines as the ability parameter. $\theta$ can be estimated by maximizing the log-likelihood through a numerical process. Subsequently, $\theta$ is sorted and can be regarded as the latest priority function $p$ in coarse-grained granularity:
\begin{equation}
    p = \operatorname{argsort}(\hat{\theta}), \qquad
    \hat{\theta}_1 > \hat{\theta}_2 > \cdots > \hat{\theta}_k.
\end{equation}
We undertake significance testing in $\theta$ to ascertain whether fine-grained experience is necessary. A one-sided Wald test  \citep{wald1943tests} is performed on the Top-ranked candidates $\theta$ at a significance level of $\alpha$. The ability gap between two candidates is measured by $\hat{d}_{ij} = \hat{\theta}_i - \hat{\theta}_j$. We regard as significant if:
\begin{equation}
\hat{d}_{ij} - z_{\alpha}\, SE(\hat{d}_{ij}) \ge 0,
\end{equation}
where $SE$ denotes the standard error of $\theta$. If the condition is met, no finer experience is demanded. Otherwise, additional fine-grained experience modeling is triggered.

In the insight level, derived from the latest estimated ability parameter $\theta$, the pairwise win/loss/tie relations can be deduced using ~\eqref{eq:btd_model}. Following this procedure, the deduced relations are then concatenated into textual form and combined with a guidance prompt. This enables LLM to distill textual experience, and it will be regarded as the latest priority function $p$ in insight granularity:
\begin{equation}
    p = \operatorname{LLM}\!\left(\text{Guidance}, \text{Concate}(P(i \succ j), P(i = j))\right).
\end{equation}

In the fine-grained level, experiences necessitate a focus on the degradation pattern. We align with the previous definition of $P$, which anchors similar degradation patterns. However, it should be noted that the categories of $P$ are not predefined in advance, and thus require a learning mechanism. A useful $P$ must satisfy the following two requirements:
\begin{itemize}
    \item \textbf{Semantic consistency} $C_s$: The images $\{I_D^{(i)} \}_{i=1}^m$ in $P$ should be close in the semantic space to form a compact feature, thus facilitating retrieval by image encoder \citep{huang2026llm2clip}.
    \item \textbf{Ranking consistency} $C_r$: The priority $p$ mapping to $P$ should be statistically comparable. This ensures that the priority $p$ of a similar $P$ in the experience pool can be reused.
\end{itemize}
In accordance with these, we design an agent-driven pattern profile learning mechanism to transform a dual-objective mathematical problem via inter-agent policy optimization. 

Using cached $\operatorname{rank}_i$ in the latest $B$ rounds, the system performs learning in a mini-batch manner. For each degraded image $I_D^i$ in the small batch, we first generate a degradation description $\text{desc}_i$ by MLLM. Subsequently, the multi-agent debate (MAD) \citep{chanchateval, smit2024should} paradigm is employed to partition the current small batch into a new pattern profile:
\begin{equation}
    P_{\text{new}} =
    \operatorname{MAD}(\text{role}, \text{action}, \mathcal{C}),
    \quad
    \mathcal{C} = \{ C_s(\text{desc}),\; C_r(\text{rank}) \},
\end{equation}
where \textit{role} specifies the agent's designated roles included in the system, \textit{action} encapsulates some predefined executable operations with dual-modality, and $\mathcal{C}(\cdot)$ denotes the dual-consistency requirements mentioned above, which are achieved as the belief \citep{rao1995bdi} for each agent.
\definecolor{top1}{RGB}{255, 230, 230} 
\definecolor{top2}{RGB}{230, 240, 255} 
\definecolor{top3}{RGB}{235, 250, 235} 
\newcommand{\first}[1]{\colorbox{top1}{\textbf{#1}}}
\newcommand{\second}[1]{\colorbox{top2}{#1}}
\newcommand{\third}[1]{\colorbox{top3}{#1}}
\begin{table}[t]
  \centering
  \scriptsize
  \caption{Quantitative comparison under different degradation settings (Groups A–C) in two \textit{Preference} (Fidelity and Perception). We compare methods within two distinct tool settings in accordance with the baseline methods for fair comparison: \textbf{Setting I} (AgenticIR-based) and \textbf{Setting II} (4KAgent-based). Rankings (\first{First}, \second{Second}, \third{Third}) are calculated independently within each group.}
  \label{tab:mioir_main_table}
  \setlength{\tabcolsep}{3.5pt} 
  \renewcommand{\arraystretch}{1.1}

    \begin{NiceTabular}{cc l ccc c ccc}[colortbl-like]
    \toprule
    \multirow{2}{*}{\textbf{Degrad.}} & \multirow{2}{*}{\textbf{Setting}} & \multirow{2}{*}{\textbf{Method}} & 
    \multicolumn{3}{c}{\textbf{Fidelity}} & & 
    \multicolumn{3}{c}{\textbf{Perception}} \\
    \cmidrule{4-6} \cmidrule{8-10}
    & & & \textbf{PSNR} $\uparrow$ & \textbf{SSIM} $\uparrow$ & \textbf{LPIPS} $\downarrow$ & & \textbf{MANIQA} $\uparrow$ & \textbf{CLIP-IQA} $\uparrow$ & \textbf{MUSIQ} $\uparrow$ \\
    \midrule
    
    \Block{12-1}{Group A} 
    & \Block{9-1}{Setting I}
    & AirNet      & 19.13 & 0.6019 & 0.4283 && 0.2581 & 0.3930 & 42.46 \\
    & & PromptIR   & 20.06 & 0.6088 & 0.4127 && 0.2633 & 0.4013 & 42.62 \\
    & & MiOIR      & 20.84 & 0.6558 & 0.3715 && 0.2451 & 0.3933 & 47.82 \\
    & & DA-CLIP    & 19.58 & 0.6032 & 0.4266 && 0.2418 & 0.4139 & 42.51 \\
    & & InstructIR & 18.03 & 0.5751 & 0.4429 && 0.2660 & 0.3528 & 45.77 \\
    & & AutoDIR    & 19.64 & 0.6286 & 0.3967 && 0.2500 & 0.3767 & 47.01 \\
    & & AgenticIR  & \second{21.04} & \second{0.6818} & \third{0.3148} && \third{0.3071} & \third{0.4474} & \third{56.88} \\
    & & MAIR       & \third{21.02} & \third{0.6715} & \second{0.2963} && \second{0.3330} & \second{0.4751} & \second{59.19} \\
    & & \textbf{EvoIR-Agent-I} & \first{23.49} & \first{0.7368} & \first{0.2840} && \first{0.3440} & \first{0.4889} & \first{60.10} \\
    \cmidrule{2-10}
    & \Block{3-1}{Setting II}
    & 4KAgent    & \third{21.48} & \third{0.6720} & \second{0.3019} && \second{0.3748} & \first{0.5544} & \second{63.19} \\
    & & TIR-Agent  & \second{22.07} & \second{0.6935} & \first{0.2874} && \first{0.3907} & \second{0.5454} & \first{63.69} \\
    & & \textbf{EvoIR-Agent-II} & \first{23.86} & \first{0.7488} & \third{0.3085} && \third{0.3623} & \third{0.5204} & \third{61.82} \\
    \midrule
    
    \Block{12-1}{Group B} 
    & \Block{9-1}{Setting I}
    & AirNet      & 19.31 & 0.6567 & 0.3670 && 0.2882 & 0.4274 & 47.88 \\
    & & PromptIR   & 20.47 & 0.6704 & 0.3370 && 0.2893 & 0.4289 & 48.10 \\
    & & MiOIR      & \third{20.56} & 0.6905 & 0.3243 && 0.2638 & 0.4330 & 51.87 \\
    & & DA-CLIP    & 18.56 & 0.5946 & 0.4405 && 0.2435 & 0.4154 & 43.70 \\
    & & InstructIR & 18.34 & 0.6235 & 0.4072 && 0.3022 & 0.3790 & 50.94 \\
    & & AutoDIR    & 19.90 & 0.6643 & 0.3542 && 0.2534 & 0.3986 & 49.64 \\
    & & AgenticIR  & 20.55 & \second{0.7009} & \third{0.3072} && \third{0.3204} & \third{0.4648} & \third{57.57} \\
    & & MAIR       & \second{20.92} & \third{0.7004} & \second{0.2788} && \first{0.3544} & \first{0.5084} & \first{60.98} \\
    & & \textbf{EvoIR-Agent-I} & \first{23.41} & \first{0.7589} & \first{0.2753} && \second{0.3535} & \second{0.4984} & \first{60.98} \\
    \cmidrule{2-10}
    & \Block{3-1}{Setting II}
    & 4KAgent    & \third{20.95} & \third{0.6727} & \second{0.3017} && \third{0.3734} & \first{0.5505} & \second{62.69} \\
    & & TIR-Agent  & \second{22.80} & \second{0.7340} & \first{0.2725} && \first{0.3997} & \second{0.5488} & \first{65.24} \\
    & & \textbf{EvoIR-Agent-II} & \first{23.38} & \first{0.7643} & \third{0.3117} && \second{0.3735} & \third{0.5296} & \third{61.96} \\
    \midrule
    
    \Block{12-1}{Group C} 
    & \Block{9-1}{Setting I}
    & AirNet      & 17.95 & 0.5145 & 0.5782 && 0.1854 & 0.3113 & 30.12 \\
    & & PromptIR   & 18.51 & 0.5166 & 0.5756 && 0.1906 & 0.3104 & 29.71 \\
    & & MiOIR      & 15.63 & 0.4896 & 0.5376 && 0.1717 & 0.2891 & 37.95 \\
    & & DA-CLIP    & 18.53 & 0.5320 & 0.5335 && 0.1916 & 0.3476 & 33.87 \\
    & & InstructIR & 17.09 & 0.5135 & 0.5582 && 0.1732 & 0.2537 & 33.69 \\
    & & AutoDIR    & 18.61 & 0.5443 & 0.5019 && 0.2045 & 0.2939 & 37.86 \\
    & & AgenticIR  & \third{18.82} & \third{0.5474} & \third{0.4493} && \third{0.2698} & \third{0.3948} & \third{48.68} \\
    & & MAIR       & \second{19.42} & \second{0.5544} & \second{0.4142} && \second{0.2798} & \second{0.4239} & \first{51.36} \\
    & & \textbf{EvoIR-Agent-I} & \first{21.52} & \first{0.6354} & \first{0.4108} && \first{0.2954} & \first{0.4448} & \first{51.36} \\
    \cmidrule{2-10}
    & \Block{3-1}{Setting II}
    & 4KAgent    & \second{19.77} & \third{0.5629} & \second{0.4271} && \second{0.3545} & \second{0.5233} & \second{55.56} \\
    & & TIR-Agent  & \third{19.53} & \second{0.5643} & \first{0.4120} && \first{0.3739} & \first{0.5495} & \first{56.93} \\
    & & \textbf{EvoIR-Agent-II} & \first{22.12} & \first{0.6588} & \third{0.4528} && \third{0.3321} & \third{0.4977} & \third{55.21} \\
    \bottomrule
  \end{NiceTabular}
  \vspace{-10pt}
\end{table}

After partitioning $P_{\text{new}}$, we further optimize and iterate on the existing $P_{\text{old}}$ through a iteration process between a soft updates to keep up with $C_s$ driven by an agent and a hard constraint to keep up with $C_r$ by $\text{Spearman} \rho$ \citep{spearman1961proof}:
\begin{equation}
    P_{\text{new}}^{t+1}
    \leftarrow
    \mathrm{Iter}\big(P_{\text{new}}^t, P_{\text{old}}^t; \text{Agent(func)}, \rho\big)
    \quad
    \rho(\text{rank}_i,\text{rank}_j) = 1 - \frac{6\sum_{i=1}^{n} \Delta_i^2}{n(n^2-1)},
\end{equation}
where \textit{func} denotes the operation functions on $P$ for agent, which includes the following meta operation: \text{add}, \text{replace}, \text{update} and \text{delete}. Through negotiation, the iteration process reaches a consensus on appending $\{I_D^{(i)} \}_{i=1}^m$ and rectifying the $text$ representation of $P$. 

Following multiple mini-batch learning steps regularized by $C_s$ and $C_r$, the information within $P$ stabilizes. Consequently, the historically cached ranks associated with the images clustered in the specific $P$ also converge. By leveraging these stable ranks, we derive the final priority function $p$ bound to $P$, which functions as the priority in fine-grained granularity:
\begin{equation}
    p = \operatorname{Stabilize}\big({\text{rank}^{t}}\big).
\end{equation}

\begin{figure}[t]
    \centering
    \begin{minipage}[c]{0.63\textwidth}
        \captionof{table}{Quantitative comparison on real-world settings from the FoundIR in two \textit{Preference} (Fidelity and Perception). Rankings are highlighted with color (\first{First}, \second{Second}, and \third{Third}).}
        \label{tab:foundir_main_table}
        \resizebox{\textwidth}{!}{%
            \setlength{\tabcolsep}{3.5pt}
            \renewcommand{\arraystretch}{1.08}
            \begin{tabular}{l ccc ccc c}
            \toprule
            \multirow{2}{*}{\textbf{Methods}} & 
            \multicolumn{3}{c}{\textbf{Fidelity}} & 
            \multicolumn{3}{c}{\textbf{Perception}} & 
            \multirow{2}{*}{\textbf{Invoc. $\downarrow$}} \\
            \cmidrule(lr){2-4} \cmidrule(lr){5-7} 
            & \textbf{PSNR} $\uparrow$ & \textbf{SSIM} $\uparrow$ & \textbf{LPIPS} $\downarrow$ & \textbf{MANIQA} $\uparrow$ & \textbf{CLIP-IQA} $\uparrow$ & \textbf{MUSIQ} $\uparrow$ & \\
            \midrule
            AgenticIR & \third{19.14} & \third{0.6283} & \third{0.3770} & \third{0.2666} & \third{0.3692} & \third{52.52} & \second{5.95} \\
            4KAgent & \second{19.76} & \second{0.7071} & \second{0.3518} & \second{0.2969} & \first{0.3873} & \second{57.35} & \third{10.89} \\
            \textbf{EvoIR-Agent} & \first{20.05} & \first{0.7101} & \first{0.3279} & \first{0.3340} & \second{0.3832} & \first{57.51} & \first{4.23} \\
            \bottomrule
            \end{tabular}%
        }
    \end{minipage}%
    \hfill
    \begin{minipage}[c]{0.34\textwidth}
        \vspace{0.1cm} 
        \centering
        \includegraphics[width=\textwidth]{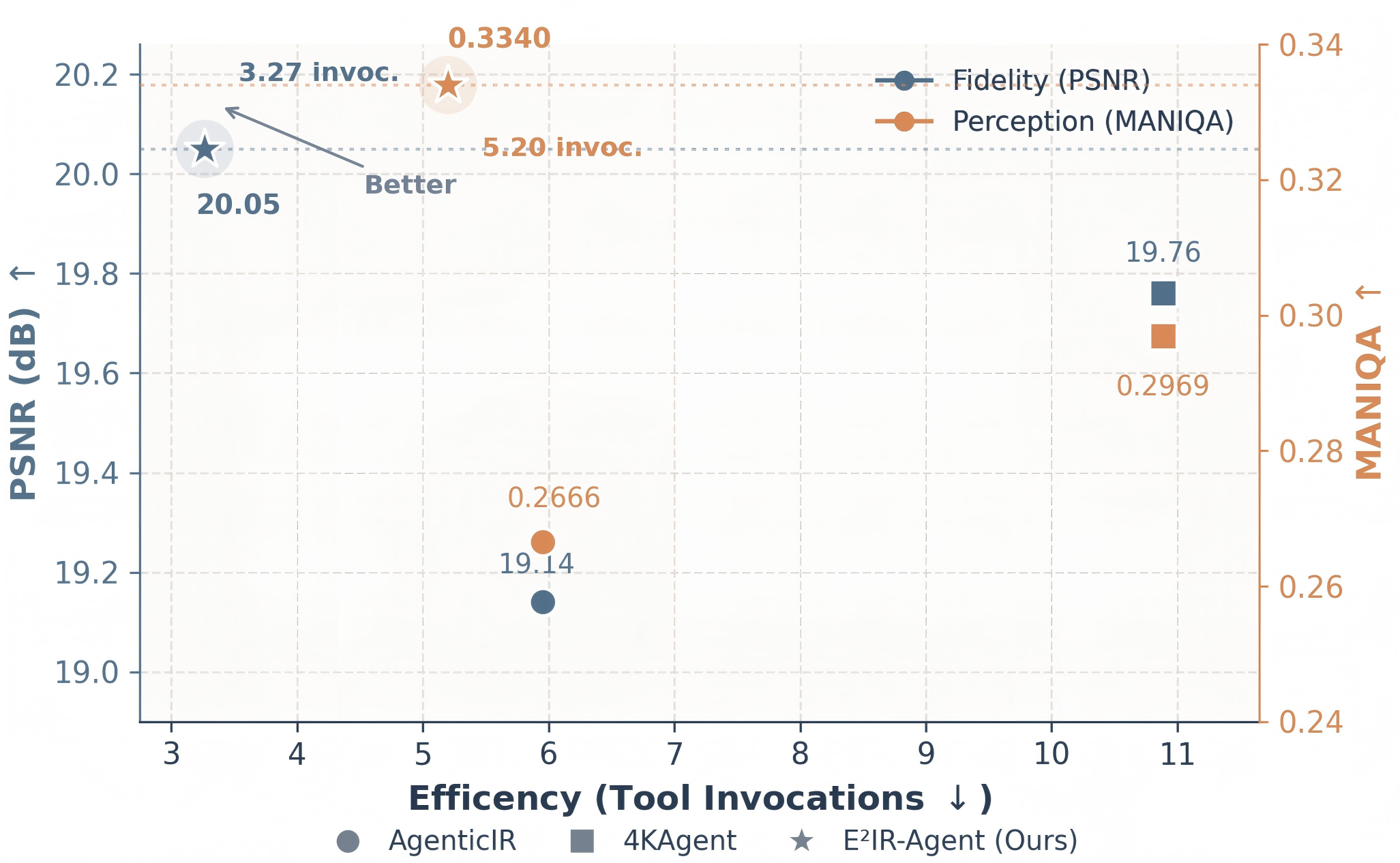}
        \vspace{-15pt}
        \captionof{figure}{Performance vs. Efficiency trade-off visualization.}
        \label{fig:foundir_main_table}
    \end{minipage}
\end{figure}

\begin{figure}[t]
  \centering
  \includegraphics[width=\textwidth]{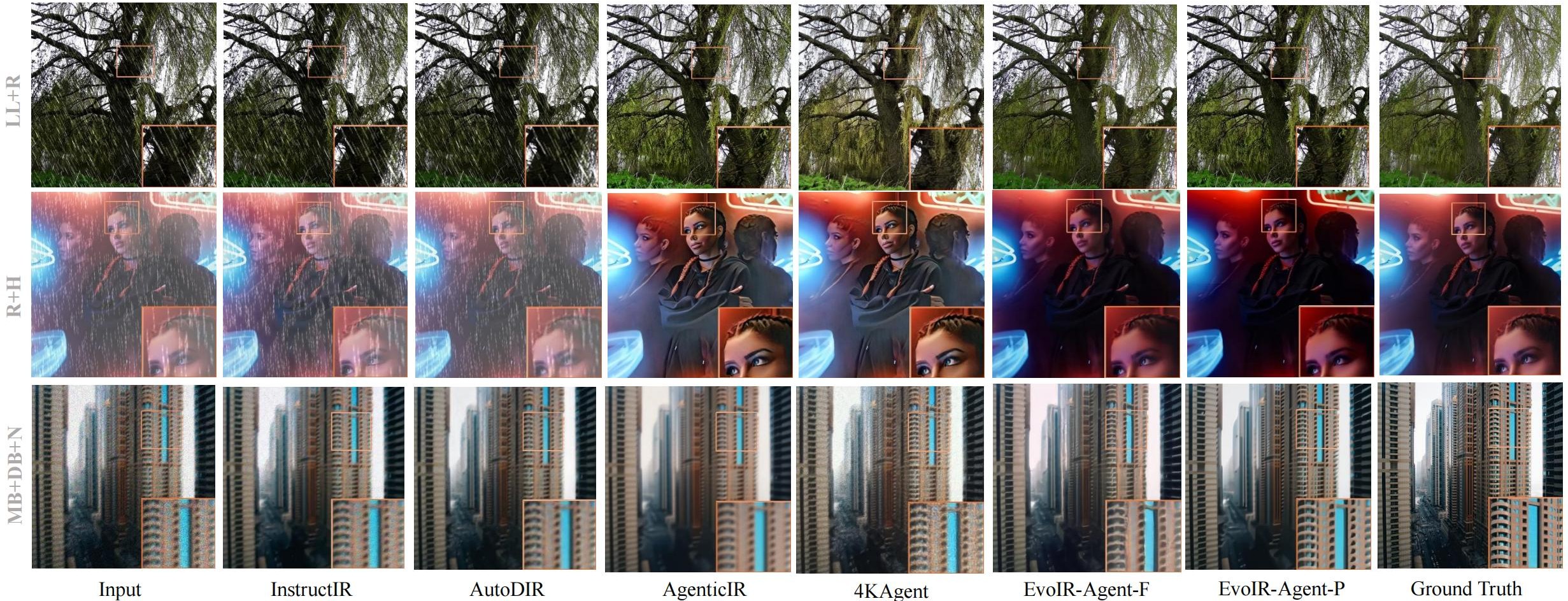}
  \caption{Visual comparison examples on MiO100 dataset. \textbf{EvoIR-Agent-F} denoted the fidelity \textit{Preference} requirement, while \textbf{EvoIR-Agent-P} denoted the perception \textit{Preference} requirement.}
  \label{fig:mioir_main_table}
\end{figure}

\section{Experiments}
\subsection{Experimental Setup}
\textbf{Test Datasets and Tool Settings.} To verify the applicability of \textbf{EvoIR-Agent}, a comprehensive testing process is conducted using both synthetic and real-world datasets. In the case of the synthesized datasets, MiO100 is employed, which contains 1,440 low-quality images across 16 combinations of two or three types of degradations. In the case of the real-world dataset, FoundIR \citep{li2025foundir} is employed for the purpose of zero-shot generalization testing in three degradation coupling scenarios. Each scenario includes 50 real-world image pairs. The detailed list of experience evolving and inference settings can be found in Appendix \ref{appendix:experiment_detail}.

\textbf{Compared Methods.} We compare our method with six AiOIR models: AirNet \citep{li2020all}, PromptIR \citep{potlapalli2023promptir}, MiOIR \citep{kong2024towards}, DA-CLIP \citep{luo2023controlling}, InstructIR \citep{conde2024instructir}, and AutoDIR \citep{jiang2024autodir}. Furthermore, we compare with four agent-based IR methods: AgenticIR \citep{zhu2025intelligent}, MAIR \citep{jiang2025multi}, 4KAgent \citep{NEURIPS2025_f0075fe4}, and TIR-Agent \citep{zhang2026tir}. All the methods above are without a re-training process.

\textbf{Evaluation Metrics.} We evaluate the quality using both full-reference (FR) and no-reference (NR) image quality assessment (IQA) metrics. The FR metrics include PSNR, SSIM \citep{wang2004image}, and LPIPS \citep{zhang2018unreasonable}, while the NR metrics include MANIQA \citep{yang2022maniqa}, CLIP-IQA \citep{wang2023exploring}, and MUSIQ \citep{ke2021musiq}.

\subsection{Performance Comparison with Other Methods} 
\textbf{Comparison with Other Methods.} As shown in Table \ref{tab:mioir_main_table}, EvoIR-Agent significantly outperforms state-of-the-art methods in Setting I in nearly all metrics, especially leading the baselines by an average of 2.35 dB in PSNR and 0.065 in SSIM. In Setting II, our performance is still competitive in FR metrics compared to the baselines and presents superior performance compared to Setting I, suggesting that our comparison of different tools between different baselines is fair and reasonable.

\textbf{Generalization to Real-World Scenarios.} Table \ref{tab:foundir_main_table} and Figure \ref{fig:foundir_main_table} evaluates the zero-shot generalization testing within the tool setting II. Our method consistently surpasses existing open-source IRAs in performance and efficiency, establishing a Pareto-optimal trade-off between restoration performance and inference efficiency. Specifically, it outperforms the strongest baselines by up to 0.0239 in LPIPS and 0.0371 in MANIQA while simultaneously slashing tool invocation number by 61\%.

\subsection{Ablation Study}
\begin{figure}[t]
    \centering
    
    \begin{minipage}[t]{0.48\textwidth}
        \centering
        \scriptsize 
        \setlength{\tabcolsep}{1.8pt} 
        \renewcommand{\arraystretch}{1.05}
        \begin{tabular}{l ccccc}
            \toprule
            \multirow{2}{*}{\textbf{Granularity}} & \multicolumn{2}{c}{\textbf{Fidelity}} & \multicolumn{2}{c}{\textbf{Perception}} & \textbf{Efficiency} \\
            \cmidrule(lr){2-3} \cmidrule(lr){4-5} \cmidrule(lr){6-6}
            & PSNR $\uparrow$ & SSIM $\uparrow$ & MANIQA $\uparrow$ & MUSIQ $\uparrow$ & Total Rb $\downarrow$ \\
            \midrule
            w/o experience & 21.02 & 0.7923 & \textbf{0.4081} & 63.21 & 3.74 \\
            \quad w/ insight & 21.56 & 0.8103 & 0.3933 & 64.58 & 2.61 \\
            \quad w/ coarse  & 22.20 & 0.8134 & 0.3913 & 64.55 & 1.46 \\
            \quad w/ fine    & \textbf{22.44} & \textbf{0.8165} & 0.3941 & \textbf{64.73} & \textbf{1.33} \\
            \bottomrule
        \end{tabular}
        \vspace{3pt}
        \subcaption{Ablation on experience granularity.}
        \label{subfig:ablation_granularity}
    \end{minipage}%
    \hfill
    \begin{minipage}[t]{0.48\textwidth}
        \centering
        \scriptsize
        \setlength{\tabcolsep}{1.8pt}
        \renewcommand{\arraystretch}{1.05}
        \begin{tabular}{c ccccc}
            \toprule
            \multirow{2}{*}{\textbf{Times}} & \multicolumn{2}{c}{\textbf{Fidelity}} & \multicolumn{2}{c}{\textbf{Perception}} & \textbf{Efficiency} \\
            \cmidrule(lr){2-3} \cmidrule(lr){4-5} \cmidrule(lr){6-6}
            & PSNR $\uparrow$ & SSIM $\uparrow$ & MANIQA $\uparrow$ & MUSIQ $\uparrow$ & Total Rb $\downarrow$ \\
            \midrule
            0 & 20.59 & 0.6579 & 0.2770 & 56.69 & 2.50 \\
            1 & 22.21 & 0.7346 & 0.3459 & 61.26 & 1.67 \\
            2 & \textbf{23.38} & 0.7643 & 0.3735 & 61.96 & \textbf{1.29} \\
            3 & 23.03 & 0.7539 & \textbf{0.3749} & \textbf{62.44} & 1.51 \\
            4 & 23.04 & \textbf{0.7647} & 0.3714 & 61.79 & 1.47 \\
            \bottomrule
        \end{tabular}
        \vspace{3pt}
        \subcaption{Ablation on evolving times.}
        \label{subfig:ablation_evolving_times}
    \end{minipage}
    
    \vspace{-2pt} 
    
    \begin{minipage}[b]{0.48\textwidth}
        \centering
        \includegraphics[width=\linewidth, height=4.2cm, keepaspectratio]{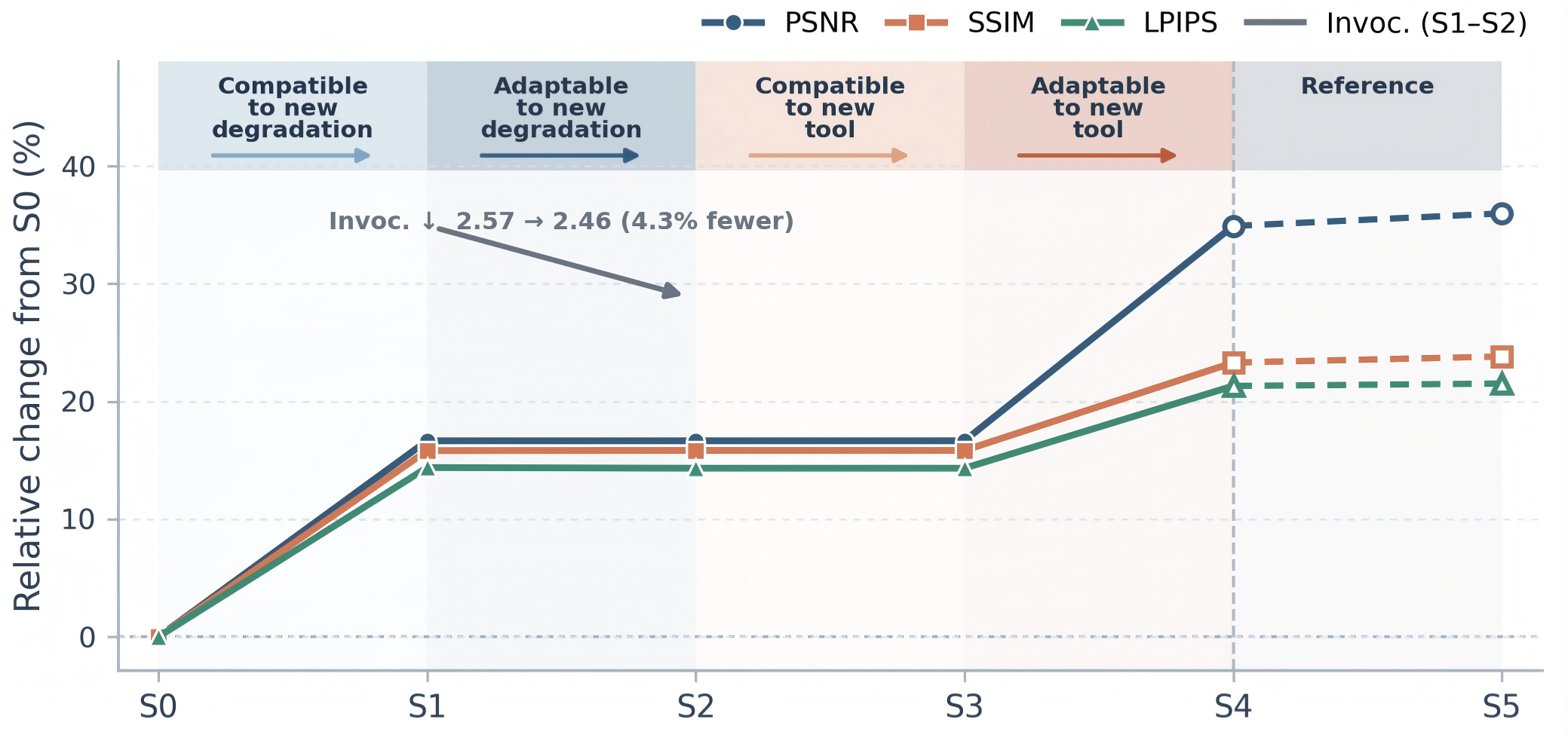}
        \vspace{-10pt}
        \subcaption{Adaptability on degradation and tool.}
        \label{subfig:demon_ada_deg_tool}
    \end{minipage}%
    \hfill
    \begin{minipage}[b]{0.48\textwidth}
        \centering
        \includegraphics[width=\linewidth, height=4.2cm, keepaspectratio]{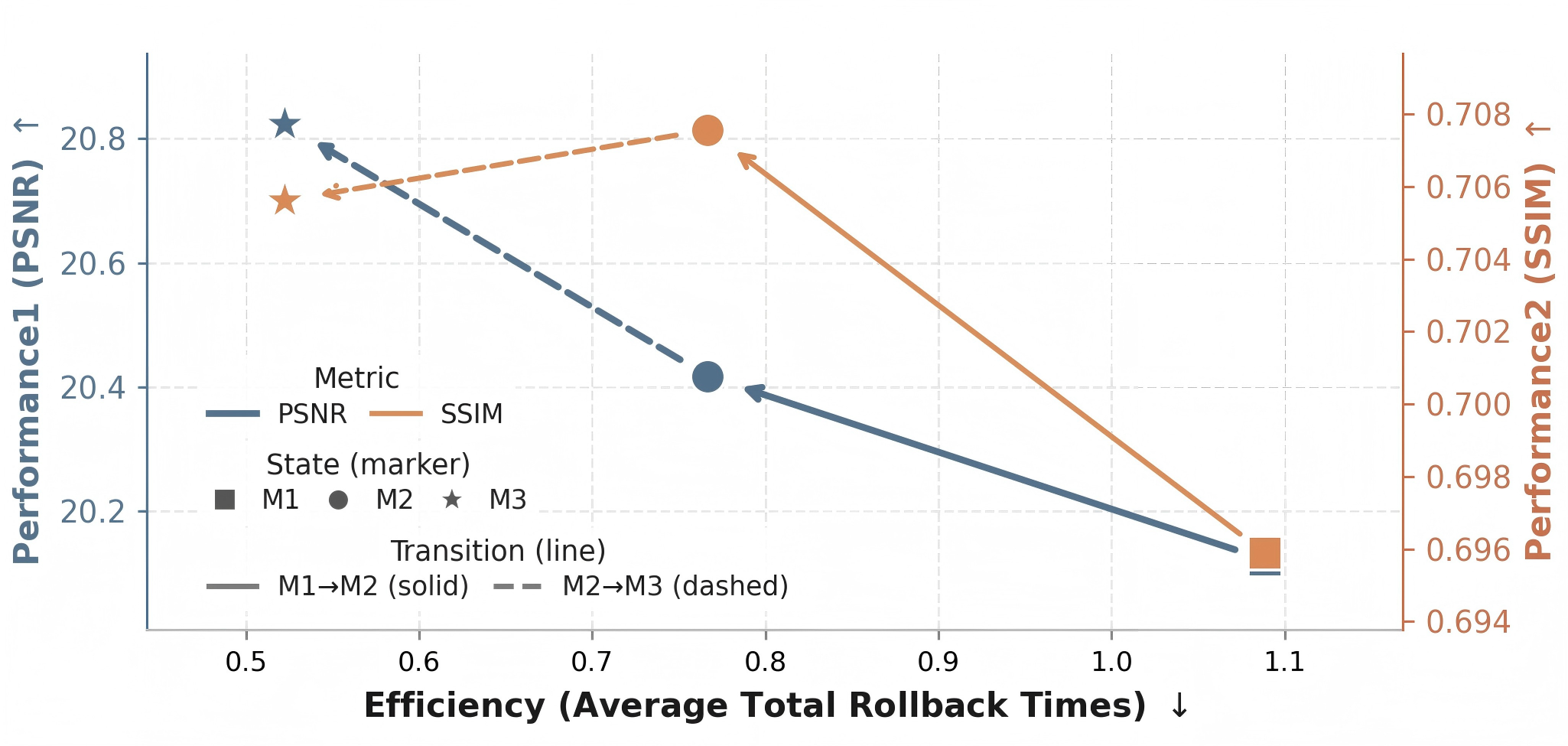} 
        \vspace{-10pt}
        \subcaption{Adaptability on data.}
        \label{subfig:demon_ada_data}
    \end{minipage}

    \caption{Ablation studies and additional experiments on core components.}
    \label{fig:main_ablation_study}
\end{figure}

\textbf{Evaluation on the hierarchical experience pool.} As shown in Table \ref{subfig:ablation_granularity}, when the granularity transitions from insight to fine, a substantial enhancement in performance and efficiency is observed in both \textit{Preference} settings. The incorporation of an experience mechanism has proven to be an effective solution to address the issue of rollback in IRAs. A comprehensive examination of the hierarchy experience trade-off is furnished in Appendix \ref{appendix:detail_ablation_experience_granularity}.

\textbf{Evaluation on the evolution times.} As shown in Table \ref{subfig:ablation_evolving_times}, $\text{Times}=2$ emerges as the optimal sweet spot between performance and efficiency when $B=25$. The self-evolving experience mechanism not only enhances performance but also reduces inference latency during evolution on the data source. A comprehensive examination of these factors is furnished in Appendix \ref{appendix:detail_ablation_evolving_times}.

\textbf{Demonstration of adaptability.} We conduct additional verification of the adaptability of our method. The self-evolving characteristic enables the system to transition from compatibility to adaptability. Figure \ref{subfig:demon_ada_deg_tool} demonstrates the ability to adapt to new degradation (from S0 to S2) and tools (from S2 to S4) in the context of continuous learning. Moreover, Figure \ref{subfig:demon_ada_data} exhibits the capacity to adapt to different data sources (from M1), facilitating the transition (M2) to the optimal theoretical state (M3). A thorough exposition of the method's adaptability is furnished in Appendix \ref{appendix:demonstration_adatability}.

\section{Conclusion}
In this paper, we address a dilemma in Image Restoration Agent (IRA) for the experience design: the limited compatibility of training-based methods versus the inference inefficiency of training-free paradigms. To resolve this, we propose \textbf{EvoIR-Agent}, which employs a hierarchical experience pool to provide coarse-to-fine guidance for three experience components. Furthermore, we introduce a self-evolving mechanism that iteratively refines this experience pool using accumulated records. This continuous evolution enables us to achieve a balance between restoration quality and efficiency while concurrently demonstrating adaptability to degradations, tools, and data sources. Extensive experiments demonstrate that we successfully bridge the gap between existing paradigms through its self-evolving, experience-driven learning capacity.

\section*{Acknowledgments}
This work was supported by the National Natural Science Foundation of China under Grant 62671633, U24A20251; Shenzhen Science and Technology Program under Grant QNXMB20250701093543046, JCYJ20230807111107015.

{\small
\bibliographystyle{unsrtnat}
\bibliography{refs}
}

\newpage
\appendix

\section{Additional Proofs}
\subsection{Experience Foundation Proving}
\label{appendix:a1_experience_proving}
\begin{figure}[htb]
  \centering
  \includegraphics[width=\textwidth]{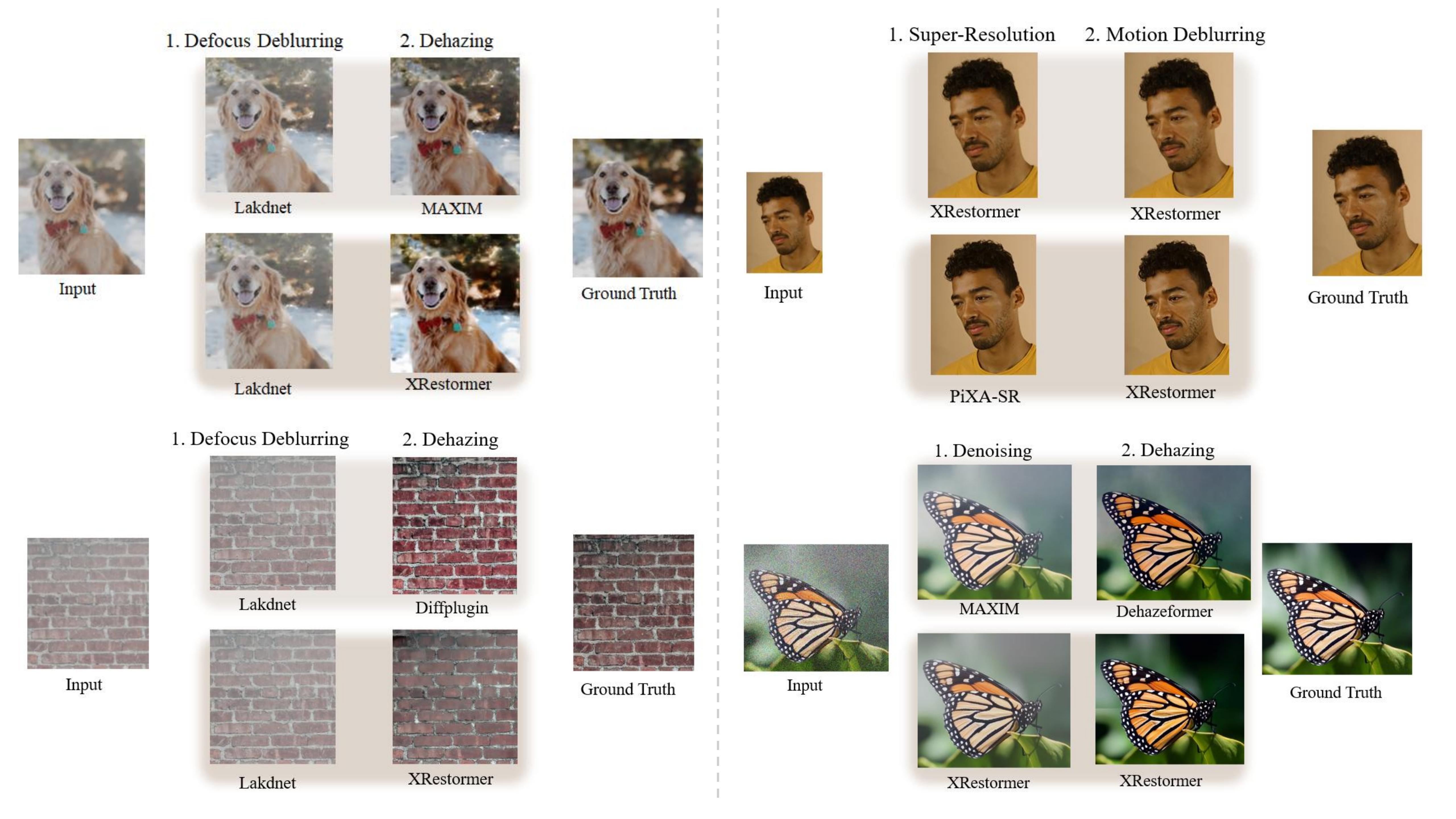}
  \caption{
  Visualization of \textit{Preference} setting. It has been demonstrated that, in instances where image restoration is achieved, the visual quality of the restored image can vary within different tools.
  }
  \label{fig:appendix_a1_experience_dimension}
\end{figure}

\textbf{Findings on 1.} Since \citep{zhu2025intelligent, NEURIPS2025_f0075fe4} utilizes a heuristic strategy to select a tool for each degradation, the restoration procedure is terminated once the degradation has been removed. Consequently, we observe that multiple iterations of the same process may yield different restored images, and they often exhibit noticeable differences in visual quality, as demonstrated in Figure \ref{fig:appendix_a1_experience_dimension}. This phenomenon motivates the introduction of \textit{Preference} as an experience component in the IRA, denoted as $Q$. This is a configurable feature for human utilization and should be objectively judged by IQA methods. We instantiate two \textit{Preference} settings (Fidelity and Perception) by using score-based IQAs. More generally, it is believed that language-based IQAs \citep{you2024depicting, wu2024q} would better align with human preference.

\textbf{Findings on 2.} Under the degradation coupling setting, jointly exploring the tools $T$ and degradation removal orders $O$ leads to a prohibitive permutational space for experience modeling. We utilize the MiOIR dataset to assess three distinct degradation coupling scenarios and fix \textit{Preference} on perception for experience evolving. This experiment is adopted to ascertain the impact of varying $p_{tool}$ values on $p_{order}$.
\begin{table}[htb]
    \centering
    \caption{Influence of $p_{tool}$ on $p_{order}$.
    ${p_{tool}}_i$ denotes the priority $i$-th for the corresponding degradation.}
    \label{tab:appendix_a1_tool_order_decoupling}
    \resizebox{\columnwidth}{!}{
    \begin{tabular}{lccccc c}
    \toprule
    \textbf{Degradation pair}
    & $\mathbf{{p_{tool}}_1+{p_{tool}}_1}$
    & $\mathbf{{p_{tool}}_1+{p_{tool}}_2}$
    & $\mathbf{{p_{tool}}_2+{p_{tool}}_1}$
    & $\mathbf{{p_{tool}}_2+{p_{tool}}_2}$
    & $\mathbf{{p_{tool}}_3+{p_{tool}}_3}$
    & \textbf{Agreement} \\
    \midrule
    Defocus blur + Haze
    & D$\rightarrow$H
    & D$\rightarrow$H
    & D$\rightarrow$H
    & D$\rightarrow$H
    & D$\rightarrow$H
    & 5/5 \\
    
    Rain + Haze
    & H$\rightarrow$R
    & R$\rightarrow$H
    & R$\rightarrow$H
    & R$\rightarrow$H
    & H$\rightarrow$R
    & 3/5 \\
    
    Motion blur + Dark
    & M$\rightarrow$B
    & M$\rightarrow$B
    & M$\rightarrow$B
    & M$\rightarrow$B
    & B$\rightarrow$M
    & 4/5 \\
    \midrule
    \multicolumn{6}{r}{\textbf{Overall modal-order agreement}}
    & \textbf{12/15 (80\%)} \\
    \bottomrule
    \end{tabular}}
\end{table}

As shown in Table \ref{tab:appendix_a1_tool_order_decoupling}, we propose to decouple two priority functions, which reduce the joint optimization from a $\prod_{d_i \in D} \mathcal{T}(d_i)$ search space down to a purely $D$-centric permutation problem. The decoupling setting enables $p_{tool}$ to be updated optionally under the degradation coupling setting.

\textbf{Findings on 3.} Subsequently, we conduct inference testing on the MiO100 \citep{kong2024towards} dataset. Our objective is to ascertain the optimal rollback strategy under the above findings.
\begin{figure}[H]
    \centering
    \begin{subfigure}[b]{0.32\textwidth}
        \centering
        \includegraphics[width=\textwidth]{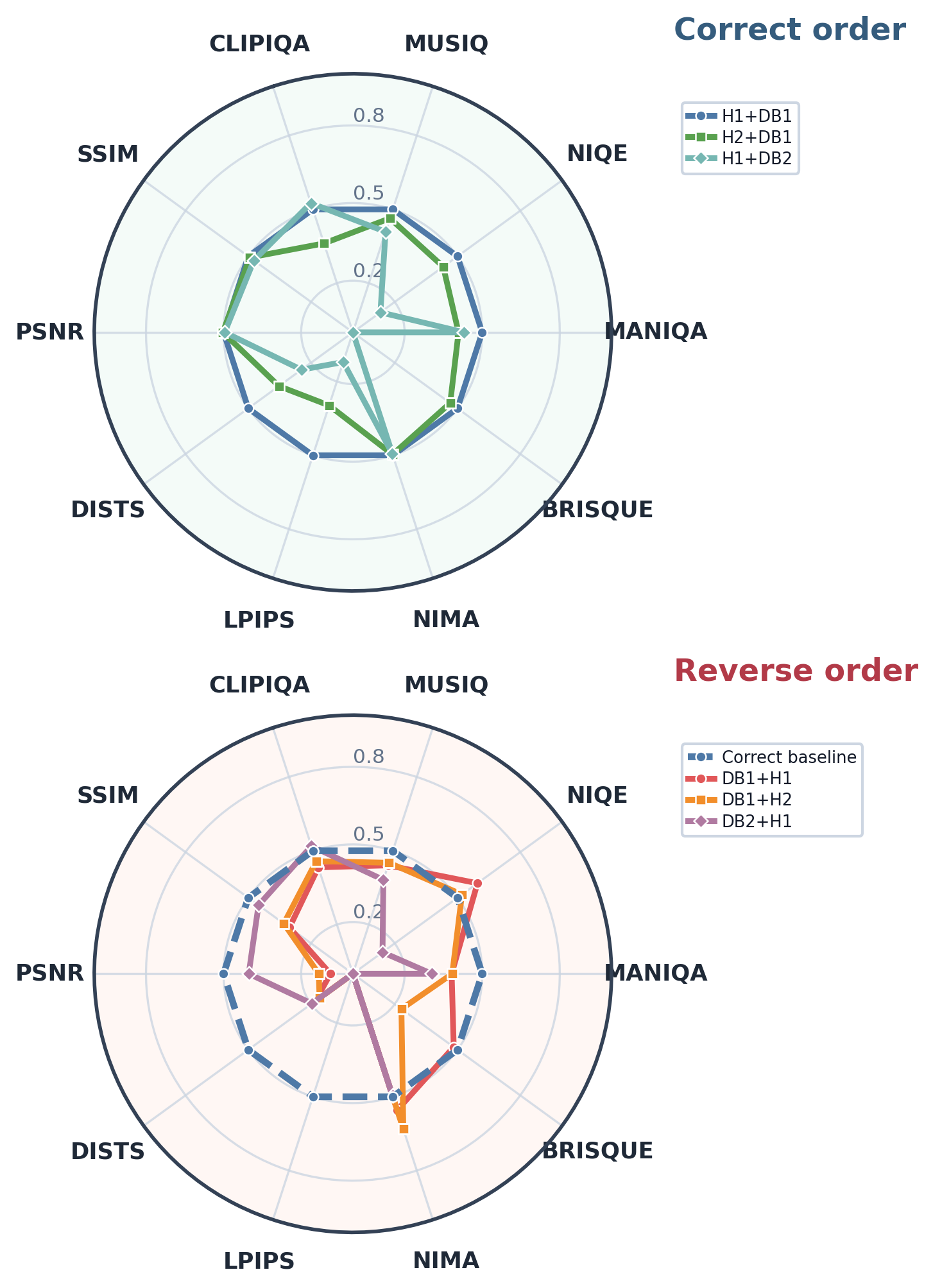} 
        \caption{Defocus Blur + Haze}
    \end{subfigure}
    \hfill 
    \begin{subfigure}[b]{0.32\textwidth}
        \centering
        \includegraphics[width=\textwidth]{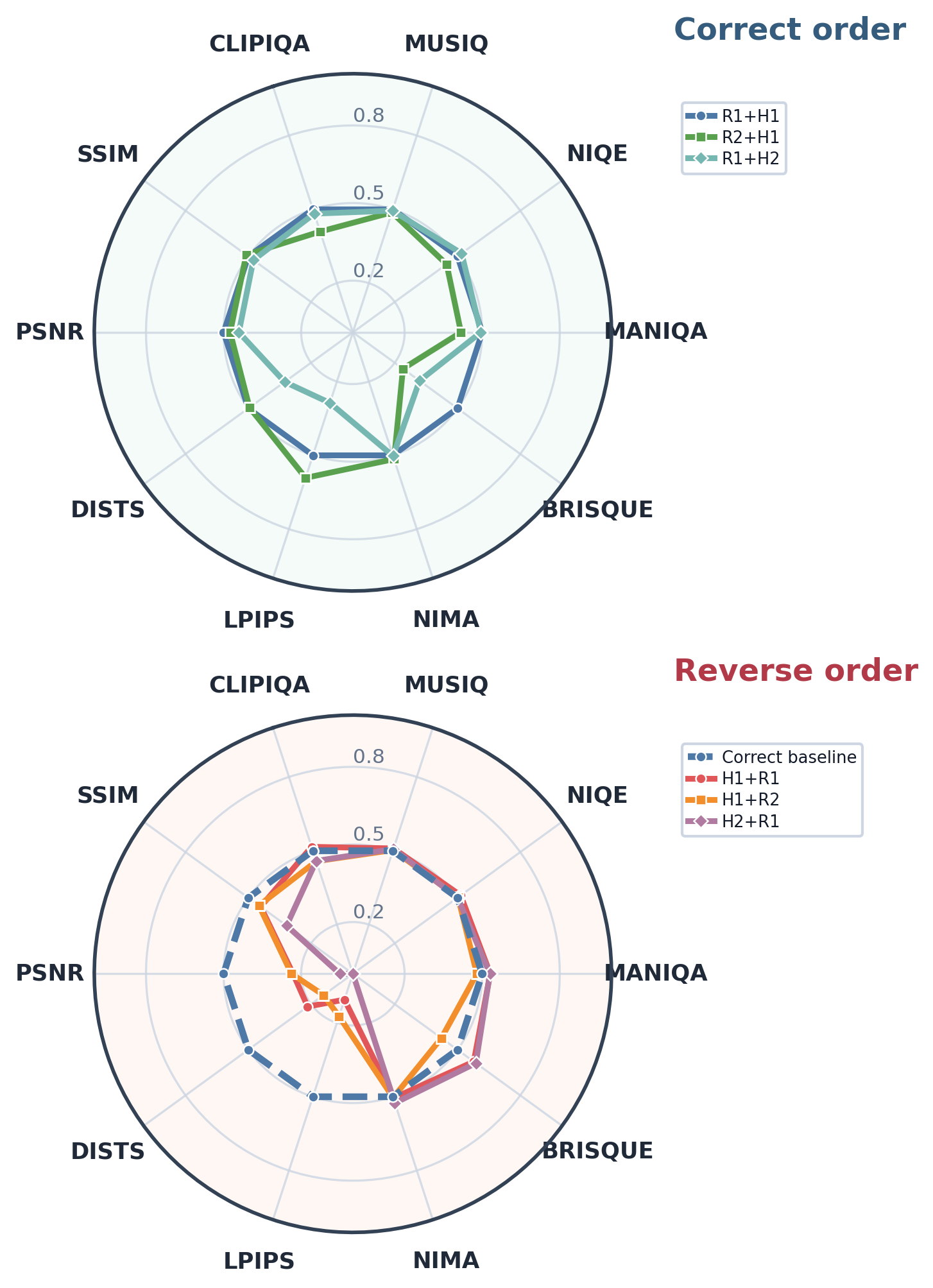} 
        \caption{Rain + Haze}
    \end{subfigure}
    \hfill
    \begin{subfigure}[b]{0.32\textwidth}
        \centering
        \includegraphics[width=\textwidth]{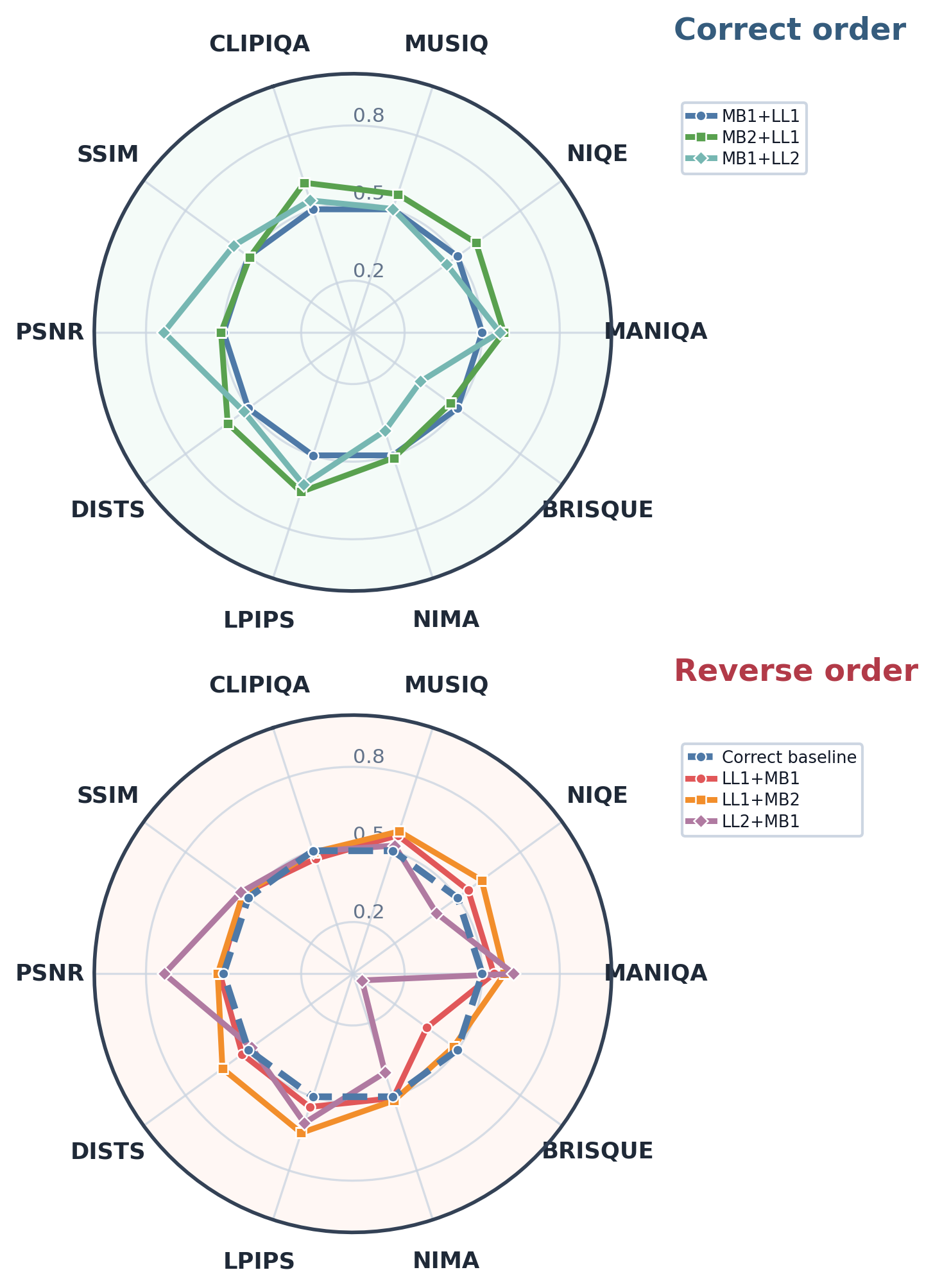} 
        \caption{Low light + Motion Blur}
    \end{subfigure}

    \caption{Influence on the rollback mechanism between $O$ and $T$. All metrics are normalized and maximized based on the optimal degradation removal order (dashed line).}
    \label{fig:appendix_a2_experience_dimension}
\end{figure}

As demonstrated in Figure \ref{fig:appendix_a2_experience_dimension}, in the event that the order is correct, the system's functionality operates within a stable boundary, and the tool selection will not result in significant fluctuations. Conversely, when the order is reversed, the functionality exhibits a pronounced decline, and the utilization of tools to mitigate this decline is not feasible. Therefore, the optimal tool is anchored in preference for $Q$ to complete the rollback in $O$ according to $p_{order}$ first and then $T$ according to $p_{tool}$.

\section{Additional Experiments}
\subsection{Demonstration from compatibility to adaptability}
\label{appendix:demonstration_adatability}
The following experiments are conducted on the FoundIR \citep{li2025foundir} dataset as in Table \ref{tab:foundir_main_table}. We partition each degradation coupling scenario into 20 images allocated for the self-evolving process and the remaining 30 images reserved for inference testing.
\subsubsection{Degradation and Tool}
Our method still follows the training-free IRAs, which are compatible with new degradations. If a tool is provided that can handle the degradation, the agentic system can be compatible without training. Furthermore, IRA can be regarded as a system of tool ensembles that is compatible with multiple new tools for one degradation.

With the assistance of the self-evolving mechanism, the evolving priority-driven mapping function $p$ allows the degradation and tool compatibility to extend to adaptability. We design six variants:

(1) \textbf{S0:} Using Tool Setting I during the evolution phase from scratch and providing only four types (total of six) of degradation tools.

(2) \textbf{S1:} Injecting two additional degradations with only one tool from Tool Setting I into S0, which presents compatibility with new degradations.

(3) \textbf{S2:} Finish the evolution process in the S1, which present the adaptability to new degradation.

(4) \textbf{S3:} Inject new tools from Tool Setting II into S2, which presents compatibility with new tools, and tools will only be added at the end of the priority function.

(5) \textbf{S4:} Finish the evolution process in the S3, which present the adaptability to new tool.

(6) \textbf{S5:} Use only Tool Setting II during the evolution phase, which is considered the reference standard.

As demonstrated by Table \ref{tab:demonstration_b1_deg_tool_adaptability} from S0 to S4, our method possesses the capability to adapt to new degradations and tools in addition to the compatibility with previous IRAs. Furthermore, the adaptability enables to approach the theoretical state in terms of performance and efficiency when compared to S5.
\begin{table}[H]
    \centering
    \small
    \setlength{\tabcolsep}{3.0pt} 
    \renewcommand{\arraystretch}{1.05} 
    \caption{A study on degradation and tool adaptability with \textbf{EvoIR-Agent}. The reference standard is highlighted in gray.}
    \label{tab:demonstration_b1_deg_tool_adaptability}
    \resizebox{\textwidth}{!}{
    \begin{tabular}{l ccccc ccccc}
        \toprule
        \multirow{3}{*}{\textbf{State}} & 
        \multicolumn{5}{c}{\textbf{Fidelity}} & 
        \multicolumn{5}{c}{\textbf{Perception}} \\
        \cmidrule(lr){2-6} \cmidrule(l){7-11}
        & \multicolumn{3}{c}{\textbf{Quality Metrics}} 
        & \multicolumn{2}{c}{\textbf{Efficiency Metrics}} 
        & \multicolumn{3}{c}{\textbf{Quality Metrics}} 
        & \multicolumn{2}{c}{\textbf{Efficiency Metrics}} \\
        \cmidrule(lr){2-4} \cmidrule(lr){5-6}
        \cmidrule(lr){7-9} \cmidrule(l){10-11}
        & PSNR $\uparrow$ & SSIM $\uparrow$ & LPIPS $\downarrow$ 
        & Invoc. $\downarrow$ & Total Rb $\downarrow$ 
        & MANIQA $\uparrow$ & CLIP-IQA $\uparrow$ & MUSIQ $\uparrow$ 
        & Invoc. $\downarrow$ & Total Rb $\downarrow$ \\
        \midrule
        $S0$ & 15.31 & 0.5699 & 0.4240 & 1.59 & 0.23 & 0.3084 & 0.3653 & 52.92 & 2.41 & 0.49 \\
        $S1$ & 17.86 & 0.6600 & 0.3630 & 2.57 & 0.39 & 0.2900 & 0.3318 & 54.79 & 3.53 & 0.71 \\
        $S2$ & 17.86 & 0.6601 & 0.3633 & 2.46 & 0.33 & 0.2918 & 0.3440 & 54.77 & 3.41 & 0.76 \\
        $S3$ & 17.86 & 0.6601 & 0.3633 & 2.46 & 0.33 & 0.2918 & 0.3440 & 54.77 & 3.41 & 0.76 \\
        $S4$ & 18.12 & 0.6764 & 0.3568 & 3.27 & 0.79 & 0.3029 & 0.3564 & 55.01 & 5.42 & 1.27 \\
        \rowcolor{gray!15}
        $S5$ & 20.66 & 0.7027 & 0.3336 & 3.10 & 0.83 & 0.3028 & 0.3753 & 56.79 & 3.53 & 0.76 \\
        \bottomrule
    \end{tabular}
    }
\end{table}

\begin{figure}[H]
    \centering
    \begin{minipage}{0.5\textwidth}
        \centering
        \includegraphics[width=1.0\linewidth]{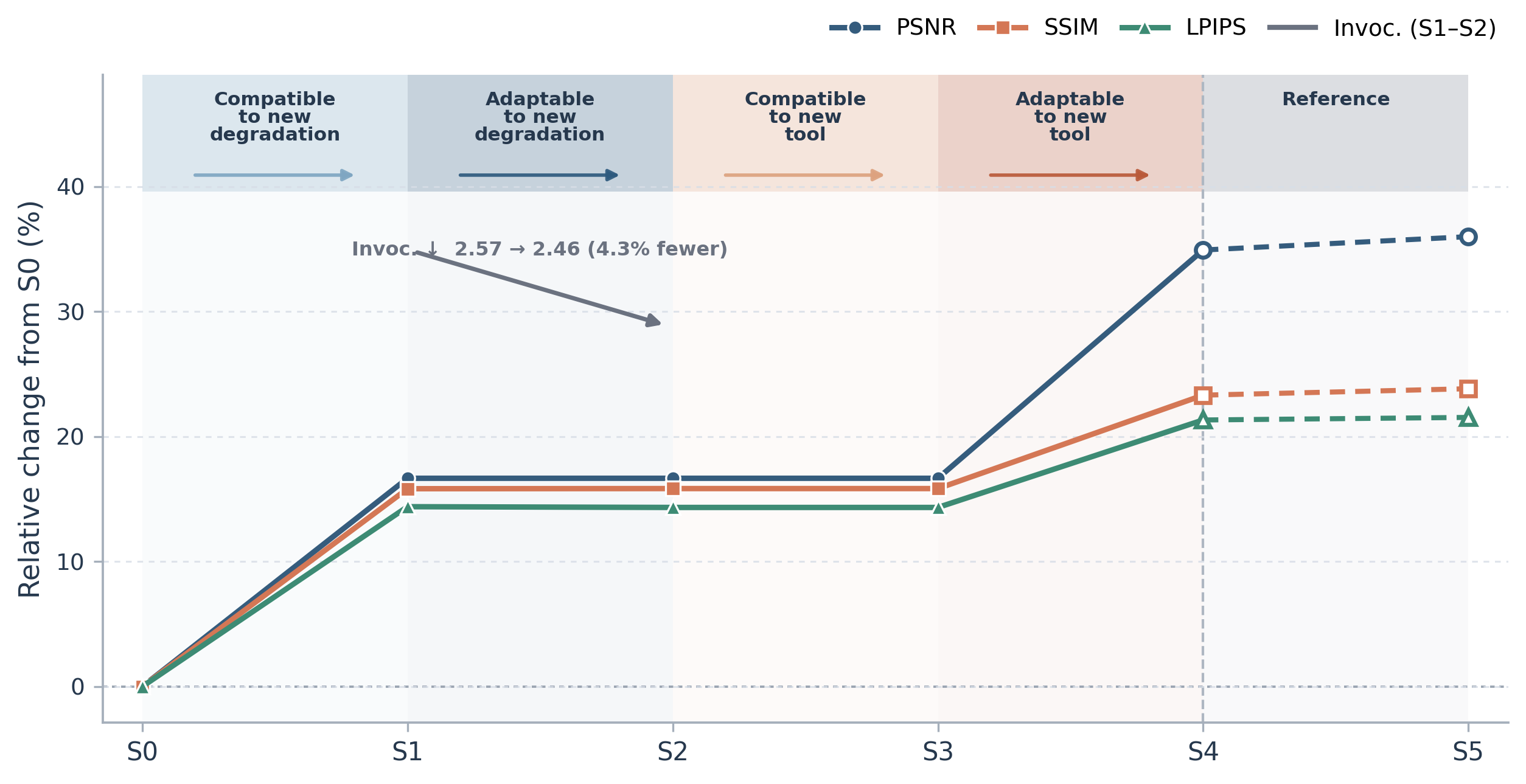}
        \subcaption{Fidelity}
    \end{minipage}
    \hfill 
    \begin{minipage}{0.5\textwidth}
        \centering
        \includegraphics[width=1.0\linewidth]{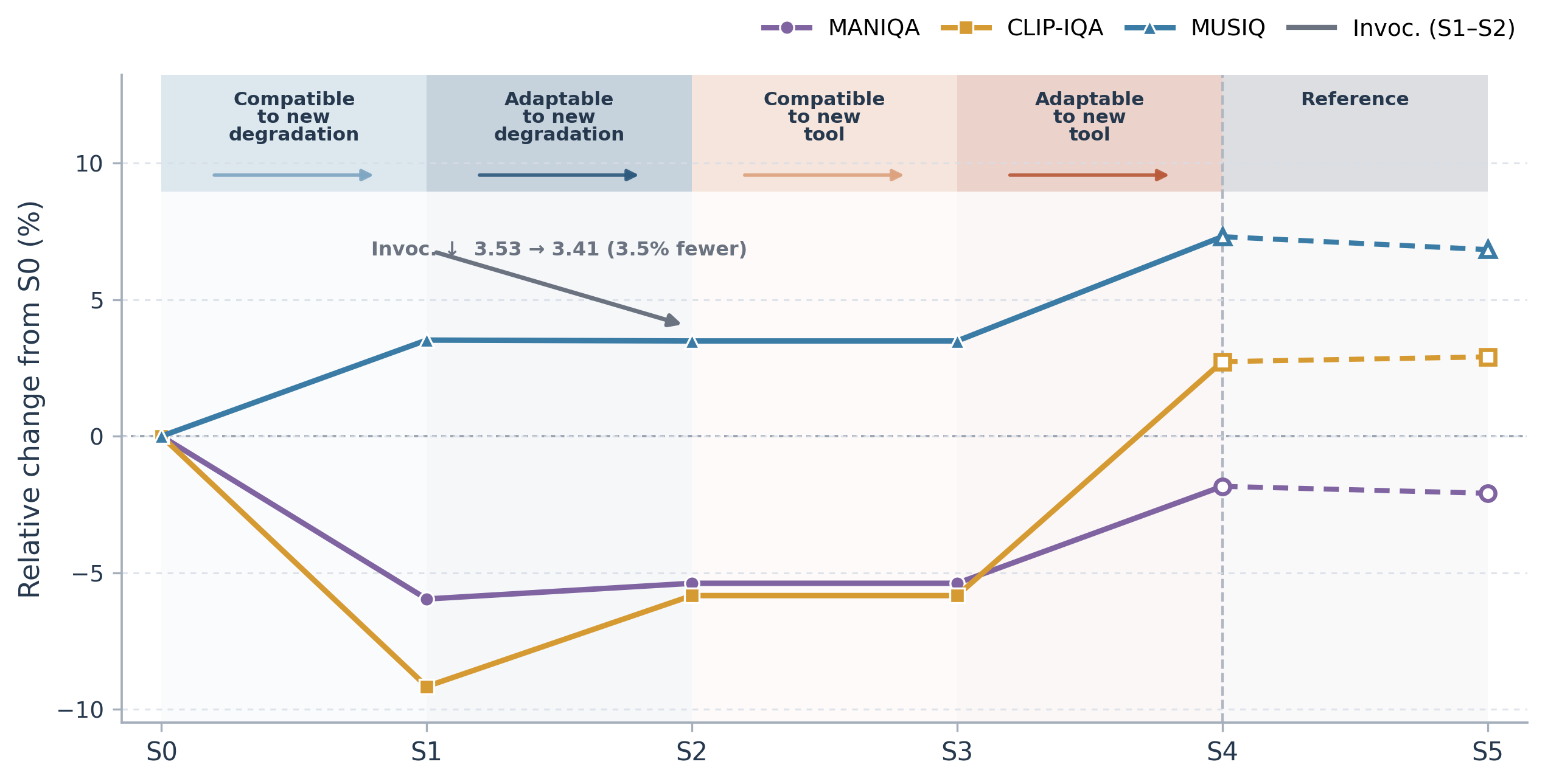}
        \subcaption{Perception}
    \end{minipage}
    \caption{Visualization of the adaptability on degradation and tool.}
    \label{fig:appendix_b1_adaptability_deg_tool}
\end{figure}
Furthermore, our method inherently supports the retirement of both degradation types and tools. 

Specifically, we adopt a soft-registration mechanism for different degradations, ensuring that inactive degradations are not loaded into GPU memory. Moreover, experiences of distinct degradations are stored independently within the experience pool, granting them seamless eliminability. 

As for tools, benefiting from the decoupling of $p_{tool}$ and $p_{order}$ in the experience representation, any retired tool is simply excised from the tool priority function $p$, with subsequent candidates shifting forward to fill the vacancy without adverse impacts.

\subsubsection{Data}
Table \ref{tab:demonstration_b1_data_adaptability} demonstrates the data source adaptability of EvoIR-Agent when confronted with entirely different data domains. To systematically analyze this adaptability, we design three variants: 

(1) \textbf{M1}: directly applies the experience gathered from the MiO100 dataset to the out-of-domain (OOD) scenario for zero-shot testing.

(2) \textbf{M2:} leverages the prior experience and continues to self-evolve using the 20 images from the OOD dataset.

(3) \textbf{M3:} initializes the self-evolving process entirely from scratch using 20 OOD images, devoid of any prior experience.

As illustrated in the experimental results, M1 and M3 provided ideal states in their respective data domains. Considering the overall performance and efficiency, M3 outperformed M1, which is in line with expectations. Meanwhile, M2 clearly demonstrates our method's ability to adapt to data, showing an overall trend moving from M1 towards M3.
\begin{table}[H]
    \centering
    \small
    \setlength{\tabcolsep}{3.0pt} 
    \renewcommand{\arraystretch}{1.05} 
    \caption{A study on data adaptability with \textbf{EvoIR-Agent}. The best results are highlighted in \textbf{bold}.}
    \label{tab:demonstration_b1_data_adaptability}
    \resizebox{\textwidth}{!}{
    \begin{tabular}{l ccccc ccccc}
        \toprule
        \multirow{3}{*}{\textbf{Methods}} &
        \multicolumn{5}{c}{\textbf{Fidelity}} &
        \multicolumn{5}{c}{\textbf{Perception}} \\
        \cmidrule(lr){2-6} \cmidrule(l){7-11}
        & \multicolumn{3}{c}{\textbf{Quality Metrics}}
        & \multicolumn{2}{c}{\textbf{Efficiency Metrics}}
        & \multicolumn{3}{c}{\textbf{Quality Metrics}}
        & \multicolumn{2}{c}{\textbf{Efficiency Metrics}} \\
        \cmidrule(lr){2-4} \cmidrule(lr){5-6}
        \cmidrule(lr){7-9} \cmidrule(l){10-11}
        & PSNR $\uparrow$ & SSIM $\uparrow$ & LPIPS $\downarrow$
        & Invoc. $\downarrow$ & Total Rb $\downarrow$
        & MANIQA $\uparrow$ & CLIP-IQA $\uparrow$ & MUSIQ $\uparrow$
        & Invoc. $\downarrow$ & Total Rb $\downarrow$ \\
        \midrule
        $M1$ & 20.12 & 0.6959 & 0.3401 & 3.50 & 1.09
        & \textbf{0.3276} & \textbf{0.3761} & 56.28 & 5.29 & 1.36 \\
        $M2$ & 20.42 & \textbf{0.7076} & 0.3425 & 3.20 & 0.77
        & 0.3178 & 0.3713 & 55.31 & 4.93 & 1.19 \\
        $M3$ & \textbf{20.82} & 0.7056 & \textbf{0.3328}
        & \textbf{2.76} & \textbf{0.52}
        & 0.3020 & 0.3759 & \textbf{56.54}
        & \textbf{3.60} & \textbf{0.80} \\
        \bottomrule
    \end{tabular}
    }
\end{table}

\subsection{Ablation Study on Hierarchical Experience Pool}
\label{appendix:detail_ablation_experience_granularity}
\begin{figure}[H]
    \centering
    \begin{minipage}{0.32\textwidth}
        \centering
        \includegraphics[width=1.0\linewidth]{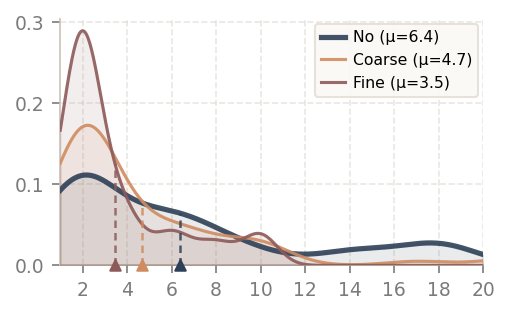} 
        \subcaption{GroupA.}
    \end{minipage}
    \hfill 
    \begin{minipage}{0.32\textwidth}
        \centering
        \includegraphics[width=1.0\linewidth]{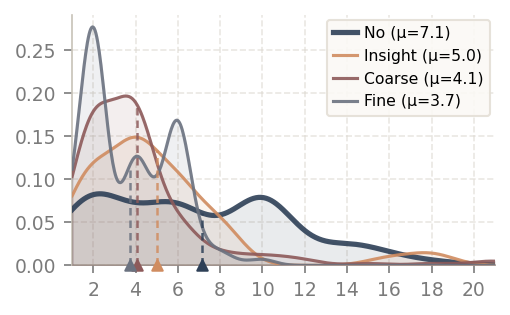} 
        \subcaption{GroupB.}
    \end{minipage}
    \hfill
    \begin{minipage}{0.32\textwidth}
        \centering
        \includegraphics[width=1.0\linewidth]{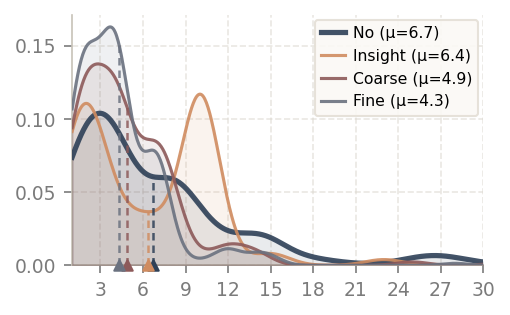}
        \subcaption{GroupC.}
    \end{minipage}
    \caption{A visualization of the frequency distribution of tool invocation numbers on different experience granularity among GroupA-C}
    \label{fig:appendix_b3_ablation_experience_granularity}
\end{figure}

Table \ref{tab:appendix_b3_ablation_experience_granularity} and Figure \ref{fig:appendix_b3_ablation_experience_granularity}  detail the ablation study of the impact of experience guidance at different granularities with $B=25$ and $Times=2$ on Group B of the MiO100. To ensure consistently processed images, this experiment is conducted on degradation types $D$, which inherently necessitates fine-grained experience. 

As demonstrated, the absence of prior experience forces the system into frequent rollback. The introduction of \textit{Insight} experience provides a global fallback mechanism, substantially reducing computational overhead. Efficiency is further enhanced as the granularity refines to \textit{Coarse} and \textit{Fine} levels, while the quality scores remain comparable.

While \textit{Fine-grained} experience offers the highest theoretical efficiency, relying on it may introduce three risks: (i) \textbf{Semantic Dependency:} Fine-grained experience relies heavily on the visual perception capability of MLLMs. If the comprehension of MLLM is limited, the \textit{Pattern Profile Learning System} (Sec. \ref{subsec:self-evolving_experience_mechanism}) may fail to generate accurate profiles. (ii) \textbf{Visual Ambiguity:} For degradations lacking salient visual features, the image encoder \citep{huang2026llm2clip} struggles to extract discriminative representations. This results in erroneous experience retrieval. (iii) \textbf{API Overhead:} Relying on fine-grained experience will incur an API cost, as shown in Table \ref{tab:appendix_c2_api_cost}.

Conversely, \textit{Coarse-grained} experience provides a more robust mapping by BTD statistically. Therefore, our hierarchical design achieves an optimal equilibrium: \textit{Coarse-grained} experience ensures generalizable stability, while \textit{Fine-grained} experience drives peak inference efficiency.
\begin{table}[htb]
    \centering
    \caption{Detailed ablation study on experience granularity on MiO100 dataset with \textbf{EvoIR-Agent}.}
    \label{tab:appendix_b3_ablation_experience_granularity}
    \resizebox{\textwidth}{!}{
    \renewcommand{\arraystretch}{1.1}
    \begin{tabular}{ll cccccc | cccc} 
        \toprule
        \multirow{2}{*}{\textbf{Degrad.}} & \multirow{2}{*}{\textbf{Methods}} & \multicolumn{6}{c}{\textbf{Quality Metrics}} & \multicolumn{4}{c}{\textbf{Efficiency Metrics}} \\
        \cmidrule(lr){3-8} \cmidrule(l){9-12} 
        & & PSNR $\uparrow$ & SSIM $\uparrow$ & LPIPS $\downarrow$ & MANIQA $\uparrow$ & CLIP-IQA $\uparrow$ & MUSIQ $\uparrow$ & Invoc. $\downarrow$ & $\mathcal{O}$-Rb $\downarrow$ & $\mathcal{T}$-Rb $\downarrow$ & Total Rb $\downarrow$ \\
        \midrule
        \multirow{3}{*}{GroupA}
        & w/o experience   & 22.12 & 0.7451 & 0.2622 & 0.3438 & 0.5035 & 61.63 & 6.38 & 1.07 & 2.20 & 3.27 \\
        & \quad w/ coarse  & 22.41 & \textbf{0.7476} & \textbf{0.2474} & 0.3503 & 0.5054 & \textbf{62.86} & 4.67 & 0.79 & 1.10 & 1.89 \\
        & \quad w/ fine    & \textbf{22.42} & 0.7464 & 0.2495 & \textbf{0.3508} & \textbf{0.5080} & 62.73 & \textbf{3.45} & \textbf{0.59} & \textbf{0.81} & \textbf{1.40} \\
        \midrule
        \multirow{4}{*}{GroupB}
        & w/o experience   & 21.02 & 0.7923 & 0.2733 & \textbf{0.4081} & \textbf{0.5858} & 63.21 & 7.13 & 0.86 & 2.88 & 3.74 \\
        & \quad w/ insight & 21.56 & 0.8103 & 0.2604 & 0.3933 & 0.5408 & 64.58 & 5.00 & 1.00 & 1.61 & 2.61 \\
        & \quad w/ coarse  & 22.20 & 0.8134 & 0.2620 & 0.3913 & 0.5317 & 64.55 & 4.08 & 0.78 & 0.68 & 1.46 \\
        & \quad w/ fine    & \textbf{22.44} & \textbf{0.8165} & \textbf{0.2576} & 0.3941 & 0.5256 & \textbf{64.73} & \textbf{3.72} & \textbf{0.67} & \textbf{0.66} & \textbf{1.33} \\
        \midrule
        \multirow{4}{*}{GroupC}
        & w/o experience   & 21.20 & 0.5745 & 0.5364 & 0.1833 & 0.2399 & 35.98 & 6.69 & 1.04 & 1.94 & 2.98 \\
        & \quad w/ insight & 21.96 & 0.5801 & 0.5770 & 0.1968 & 0.2811 & 37.48 & 6.37 & \textbf{0.62} & 2.35 & 2.97 \\
        & \quad w/ coarse  & \textbf{22.15} & \textbf{0.5943} & 0.5130 & \textbf{0.2037} & \textbf{0.2865} & 40.57 & 5.04 & 1.01 & 1.11 & 2.12 \\
        & \quad w/ fine    & \textbf{22.15} & 0.5942 & \textbf{0.5123} & 0.2031 & 0.2850 & \textbf{41.09} & \textbf{4.32} & 0.85 & \textbf{0.89} & \textbf{1.74} \\
        \bottomrule
    \end{tabular}
    }
\end{table}

\subsection{Visualization on Hierarchical Experience}
\begin{figure}[H]
    \centering
    \begin{minipage}{0.5\textwidth}
        \centering
        \includegraphics[width=1.0\linewidth]{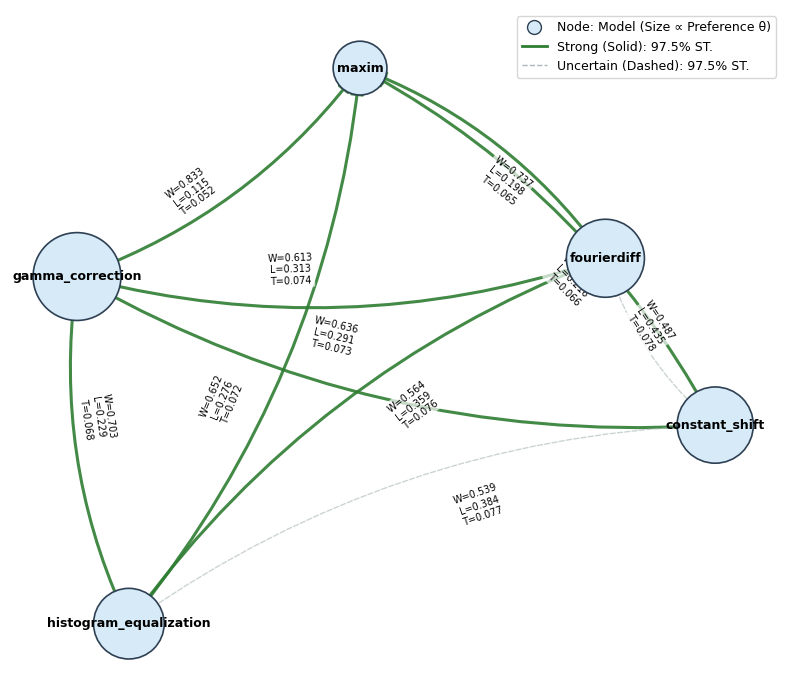}
        \subcaption{BTD Complete Graph on 'Dark'}
    \end{minipage}
    \hfill 
    \begin{minipage}{0.5\textwidth}
        \centering
        \includegraphics[width=1.0\linewidth]{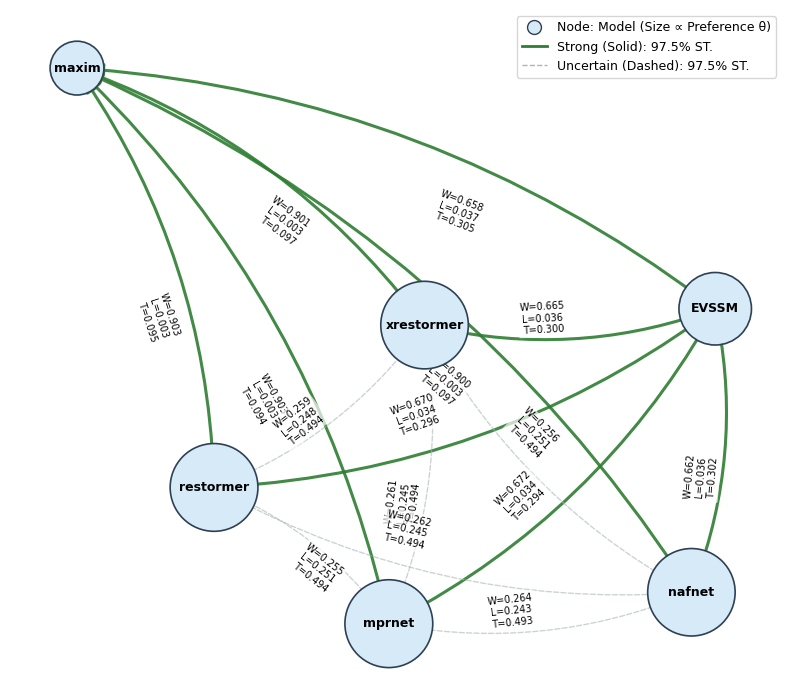}
        \subcaption{BTD Complete Graph on 'Motion Blur'}
    \end{minipage}
    \caption{Visualization of the BTD Model result in \textit{Tool Selection}. The size of the nodes corresponds to the ability parameter as estimated by the BTD model. In instances where the coarse-grained experience is uncertain (represented by a dashed line), thereby signifying that fine-grained experience must be utilized.}
    \label{fig:appendix_b4_btd_vis_tools}
\end{figure}

\begin{figure}[H]
    \centering
    \begin{minipage}{0.5\textwidth}
        \centering
        \includegraphics[width=1.0\linewidth]{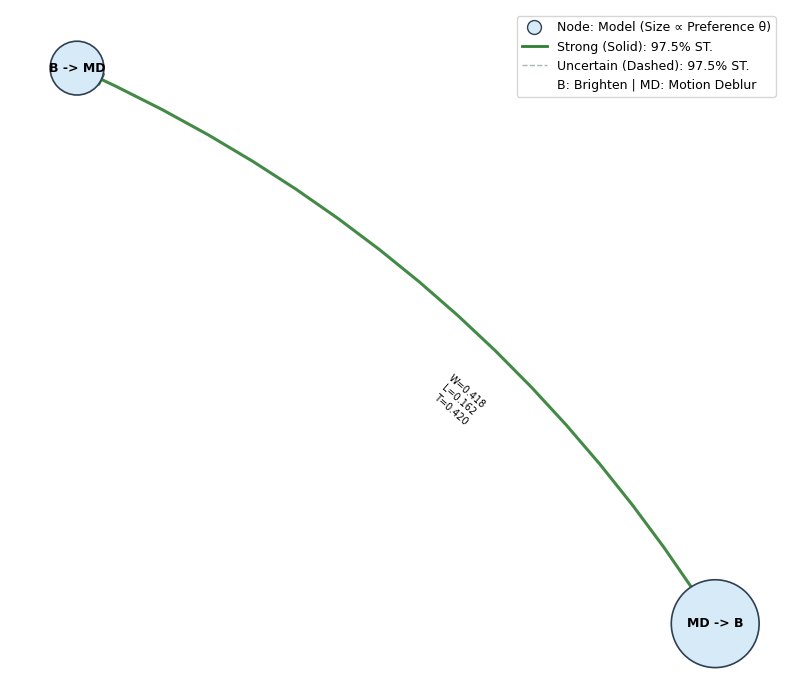}
        \subcaption{BTD Complete Graph on 'Dark+Motion Blur'}
    \end{minipage}
    \hfill 
    \begin{minipage}{0.5\textwidth}
        \centering
        \includegraphics[width=1.0\linewidth]{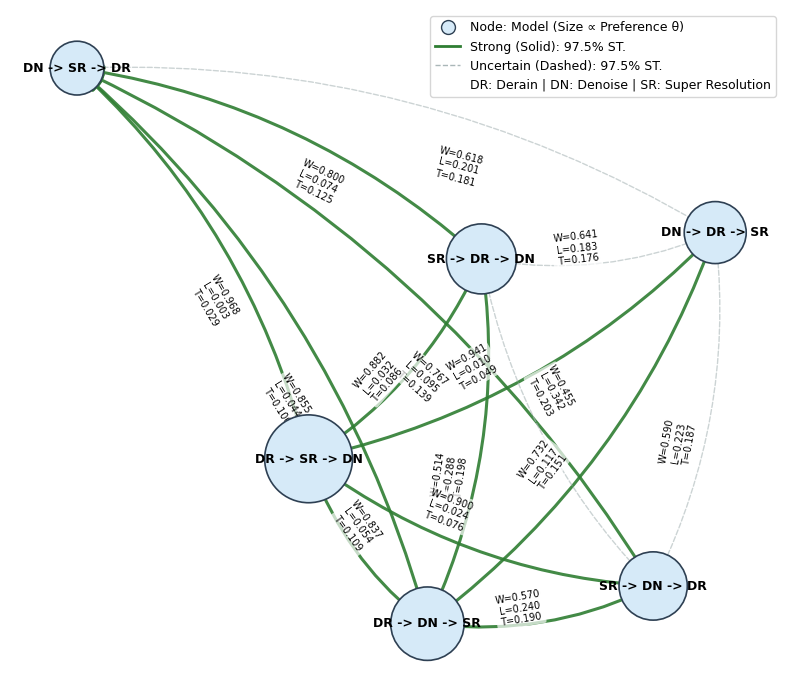}
        \subcaption{BTD Complete Graph on 'Rain+Haze+LR'}
    \end{minipage}
    \caption{Visualization of the BTD Model result in \textit{Degradation Removal Order}.}
    \label{fig:appendix_b4_btd_vis_removal_order}
\end{figure}

In this section, we will present several sample sections (insight: Table \ref{tab:appendix_b4_insight_exp_vis}; coarse-grained: Table \ref{tab:appendix_b4_coarse_exp_vis}; fine-grained: Table \ref{tab:appendix_b4_fine_exp_vis}) of the hierarchical experience pool to facilitate comprehension. Furthermore, a graph will be constructed to visualize the BTD statistical model underlying the findings, as shown in Figure \ref{fig:appendix_b4_btd_vis_tools} and Figure \ref{fig:appendix_b4_btd_vis_removal_order}.
\vspace{-10pt}
\begin{table}[H]
\centering
\caption{Experience sample of \textbf{insight}}
\begin{tabular}{>{\centering\arraybackslash}m{2cm} m{11cm}}
\hline
\textbf{Preference} & \multicolumn{1}{c}{\textbf{Experience}} \\ 
\hline
Fidelity & Here is the reference information from past trials: A reasonable overall elimination order that respects all strong constraints and most weaker ones is: JPEG artifacts → Defocus blur → Haze → Rain → Motion blur → Brightening → Denoising → Super-resolution. \\
\hline
Perception & Here is the reference information from past trials: Recommended "default" sequence, JPEG compression artifacts removal → Defocus deblurring → Dehazing → Deraining → Super-resolution → Motion deblurring → Denoising (only confidently "after JPEG" and "before brightening" \\
\hline
\label{tab:appendix_b4_insight_exp_vis}
\end{tabular}
\end{table}
\begin{table}[H]
\centering
\caption{Experience sample of \textbf{coarse-grained experience}}
\begin{tabular}{>{\centering\arraybackslash}m{2cm} >{\centering\arraybackslash}m{2cm} m{9cm}}
\hline
\textbf{Degradation} & \textbf{Preference} & \multicolumn{1}{c}{\textbf{Experience}} \\
\hline

\multirow{15}{*}{Motion Blur} 
& Fidelity &
\begin{minipage}{7cm}
\begin{lstlisting}[basicstyle=\ttfamily\small]
"degradation_type": "motion blur",
"preference": "fidelity",
"ranking": {
    "mprnet": 1,
    "restormer": 2,
    "xrestormer": 3,
    ...
\end{lstlisting}
\end{minipage} \\

& Perception &
\begin{minipage}{7cm}
\begin{lstlisting}[basicstyle=\ttfamily\small]
"degradation_type": "motion blur",
"preference": "perception",
"ranking": {
    "xrestormer": 1,
    "restormer": 2,
    "EVSSM": 3,
    ...
\end{lstlisting}
\end{minipage} \\

\hline

\multirow{12}{*}{D+M} 
& Fidelity &
\begin{minipage}{7cm}
\begin{lstlisting}[basicstyle=\ttfamily\small]
"degradation_type": "dark+motion blur",
"preference": "fidelity",
"ranking": {
    "motion blur -> dark": 1,
    "dark -> motion blur": 2}
\end{lstlisting}
\end{minipage} \\

& Perception &
\begin{minipage}{7cm}
\begin{lstlisting}[basicstyle=\ttfamily\small]
"degradation_type": "dark+motion blur",
"preference": "perception",
"ranking": {
    "dark -> motion blur": 1,
    "motion blur -> dark": 2}
\end{lstlisting}
\end{minipage} \\

\hline
\label{tab:appendix_b4_coarse_exp_vis}
\end{tabular}
\end{table}

\begin{table}[H]
\centering
\caption{Experience sample of \textbf{fine-grained experience}}
\begin{tabular}{>{\centering\arraybackslash}m{2cm} >{\centering\arraybackslash}m{2cm} m{9cm}}
\hline
\textbf{Degradation} & \textbf{Preference} & \multicolumn{1}{c}{\textbf{Experience}} \\
\hline
\multirow{8}{*}{Motion Blur} 
& \multirow{8}{*}{Fidelity} &
\begin{minipage}{7cm}
\begin{lstlisting}[basicstyle=\ttfamily\small]
{
    "exp_id": 0,
    "degradation_type": "motion blur",
    "preference": "fidelity",
    "degradation_pattern": "Significant motion...",
    "ranking": {
        "xrestormer": 1,
        "mprnet": 2,
        ...
        },
    "related_trajectory_ids": [
        0,
        6,
        ...
    ]
},
{
    "exp_id": 1,...
\end{lstlisting}
\end{minipage} \\

\hline

\multirow{12}{*}{D+M} 
& \multirow{12}{*}{Perception} &
\begin{minipage}{7cm}
\begin{lstlisting}[basicstyle=\ttfamily\small]
{
    "exp_id": 0,
    "degradation_type": "dark+motion blur",
    "preference": "perception",
    "degradation_pattern": "Significant loss...",
    "ranking": {
        "motion blur -> dark": 1,
        "dark -> motion blur": 2
    },
    "related_trajectory_ids": [
        0,
        1,
        ...
    ]
},
{
    "exp_id": 1,...
\end{lstlisting}
\end{minipage} \\

\hline
\label{tab:appendix_b4_fine_exp_vis}
\end{tabular}
\end{table}

\subsection{Ablation Study on Evolution Times}
\label{appendix:detail_ablation_evolving_times}

\begin{figure}[htb]
    \centering
    \begin{minipage}{0.31\textwidth}
        \centering
        \includegraphics[width=1.0\linewidth]{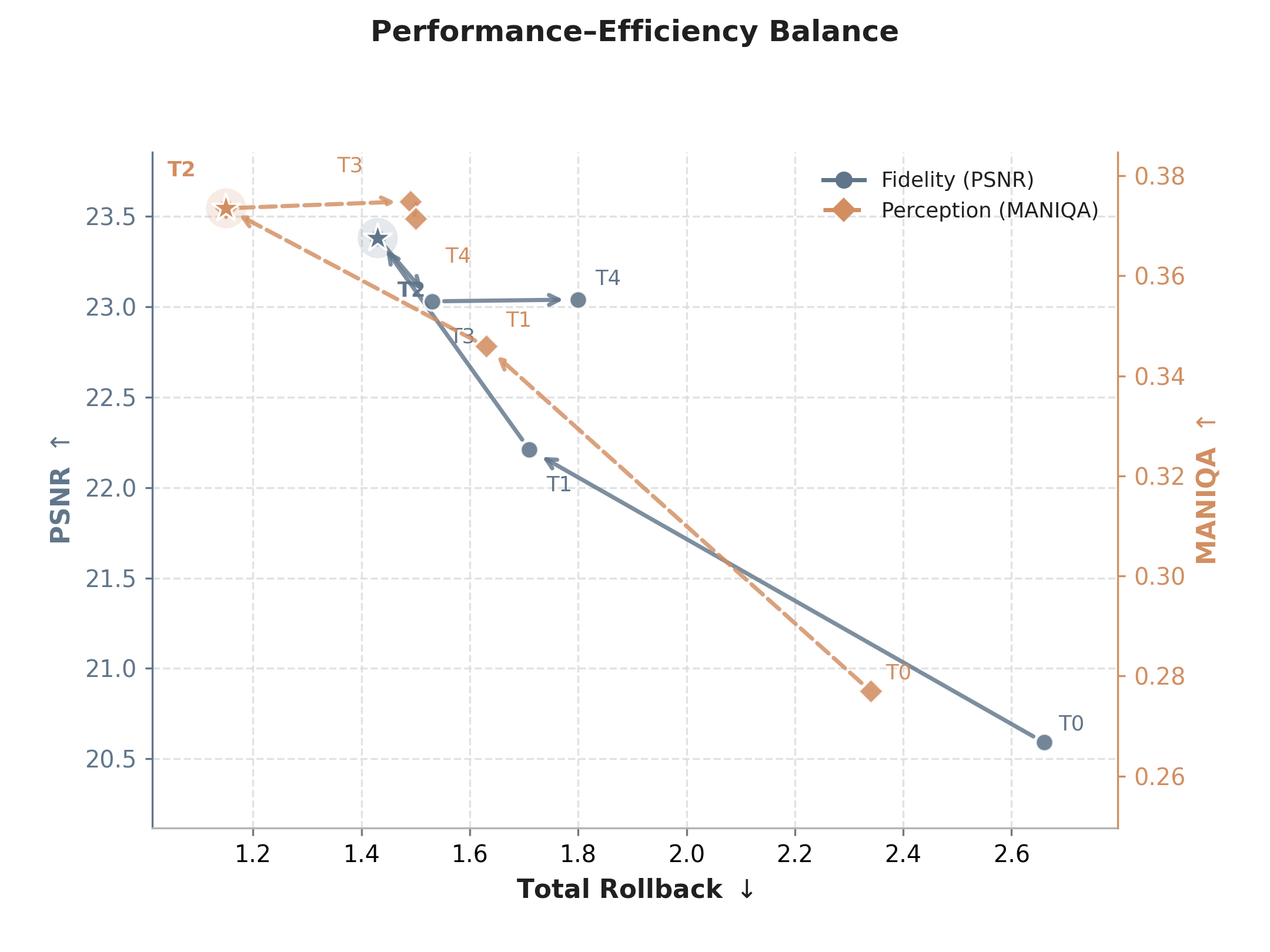} 
        \subcaption{Quality-Efficiency Trade-off.}
    \end{minipage}%
    \hfill 
    \begin{minipage}{0.31\textwidth}
        \centering
        \includegraphics[width=1.0\linewidth]{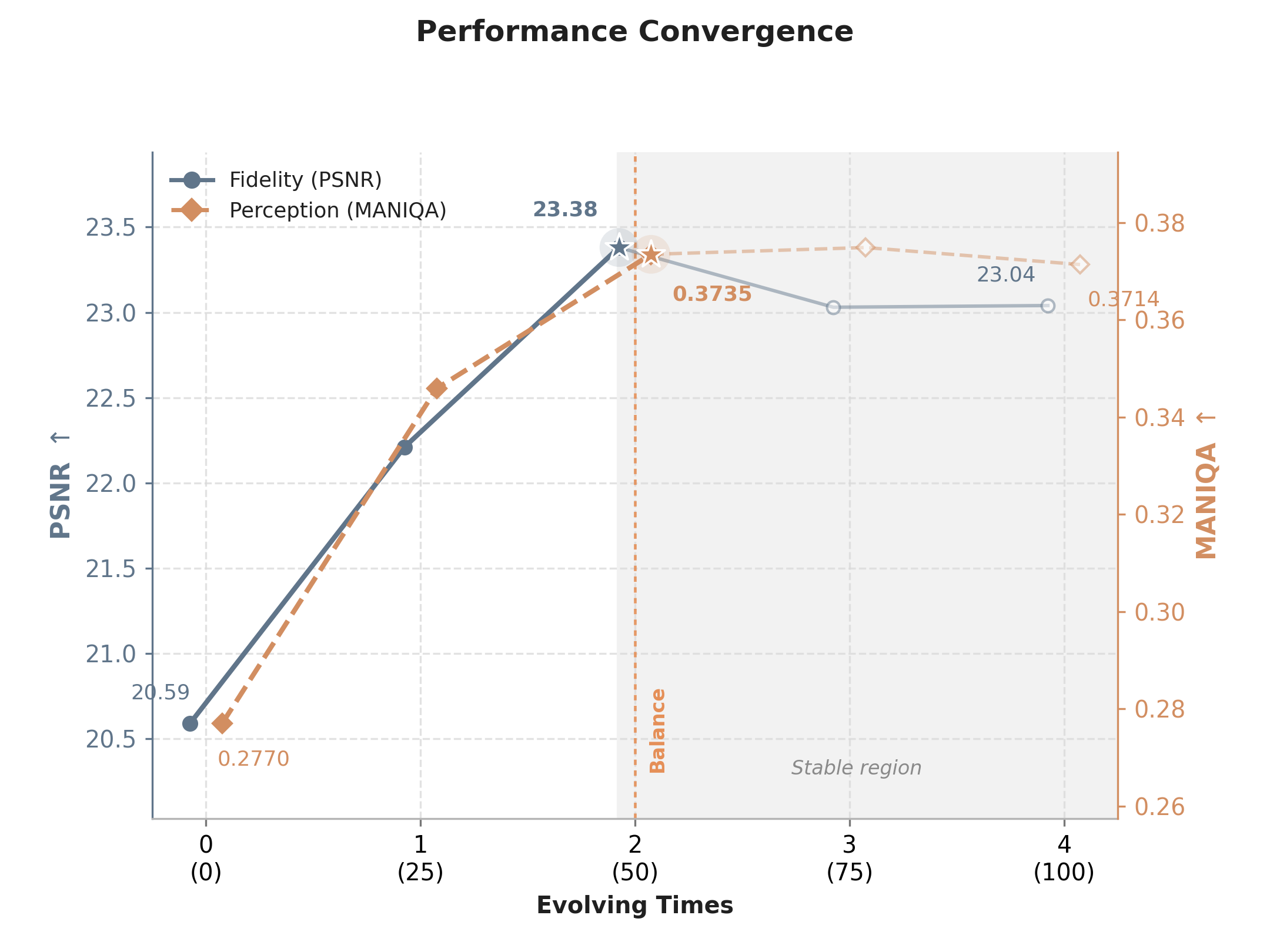} 
        \subcaption{Quality Progression.}
    \end{minipage}%
    \hfill 
    \begin{minipage}{0.31\textwidth}
        \centering
        \includegraphics[width=1.0\linewidth]{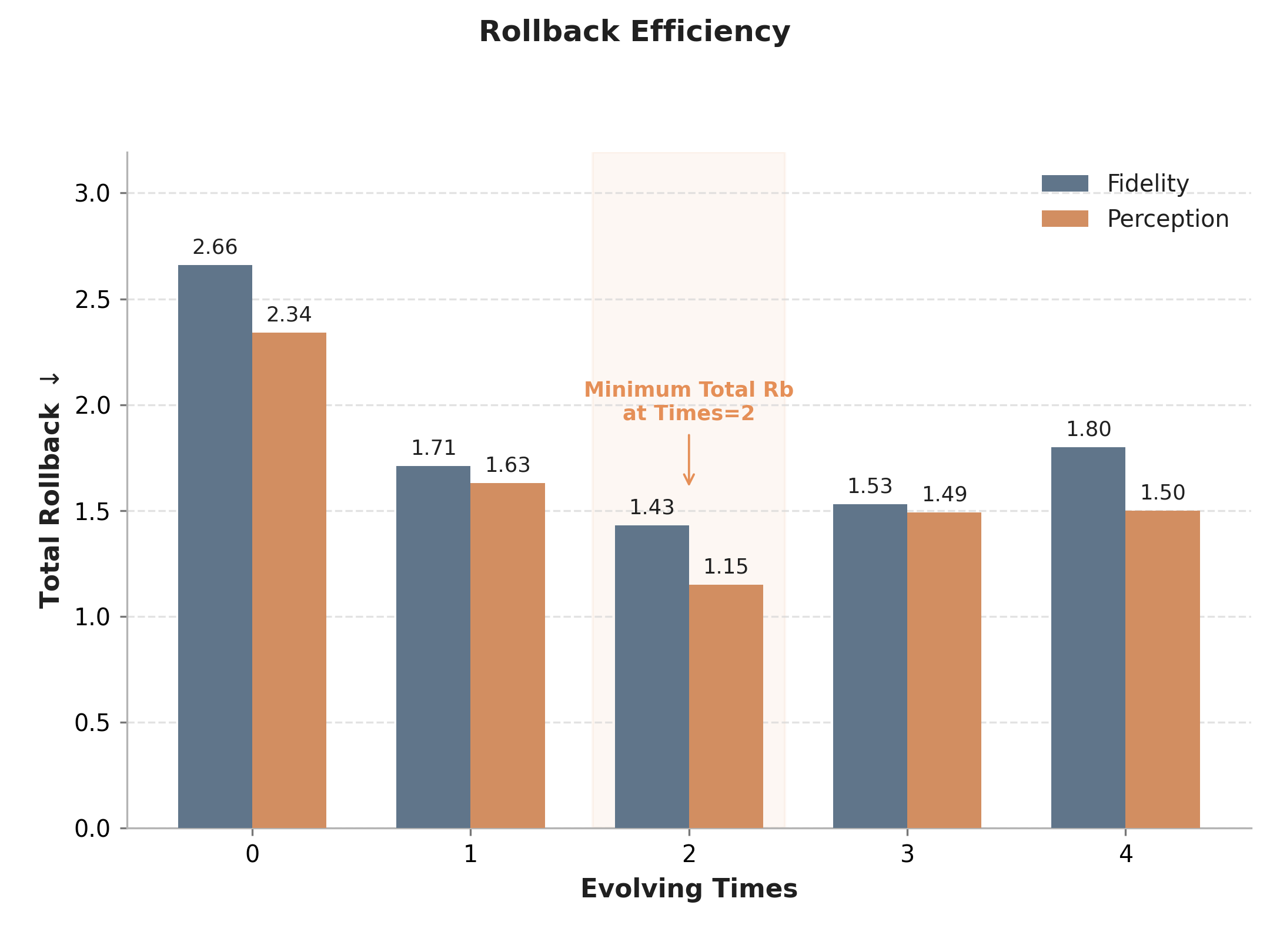}
        \subcaption{Efficiency Measurement.}
    \end{minipage}
    \caption{
        Performance-Efficiency analysis of the evolution process with batch size $B$=25. (a) Cost-quality Pareto front between performance and efficiency in both \textit{Preference}. (b) Quality progression over evolution times. (c) Efficiency measurement by the total rollback times. 
    }
    \label{fig:ablation_evolve_times_detail}
\end{figure}

Table \ref{tab:ablation_evolve_times_detail} and Figure \ref{fig:ablation_evolve_times_detail} detail the ablation study of the self-evolving experience mechanism on Group B of the MiO100 dataset with batch size $B=25$. By scanning the evolution times, we seek to investigate the optimal balance point.

Overall, $\text{Times}=2$ emerges as the optimal sweet spot between performance and efficiency. Now we detail the analysis: (i) \textbf{Cost-Quality:} As illustrated in (a), the system's evolution exhibits a convergent trajectory from the initial condition of $\text{Times}=0$ to the condition of $\text{Times}=2$, which reaches the Pareto-optimal point in both \textit{Preference} at $\text{Times}=2$ (top-left corner). However, for $\text{Times}>2$, the curve folds back towards the bottom-right, indicating potential experience redundancy. (ii) \textbf{Quality Progression:} The findings in (b) indicate that both \textit{Preference} exhibit a marked increase in growth during the initial phase ($\text{Times} < 2$) and attains its peak level of growth at the $\text{Times}=2$ mark. The plateau and slight fluctuations show diminishing returns in the subsequent phase ($\text{Times} > 2$). (iii) \textbf{Efficiency Measurement:} Furthermore, the efficiency measurement in (c) explains that the substantial decrease in total rollback times stop at $\text{Times}=2$.
\begin{table}[htb]
    \centering
    \small
    \setlength{\tabcolsep}{4.0pt} 
    \renewcommand{\arraystretch}{1.05}
    \caption{Ablation study on evolving times with \textbf{EvoIR-Agent}. The best results are highlighted in \textbf{bold}.}
    \label{tab:ablation_evolve_times_detail}
    \resizebox{\textwidth}{!}{
    \begin{tabular}{cc cccc cccc}
        \toprule
        \multirow{3}{*}{\textbf{Times}} & \multirow{3}{*}{\textbf{$\boldsymbol{B=25}$}} & 
        \multicolumn{4}{c}{\textbf{Fidelity}} & 
        \multicolumn{4}{c}{\textbf{Perception}} \\
        \cmidrule(lr){3-6} \cmidrule(l){7-10}
        & & \multicolumn{3}{c}{\textbf{Quality Metrics}} & \textbf{Efficiency Metrics} 
        & \multicolumn{3}{c}{\textbf{Quality Metrics}} & \textbf{Efficiency Metrics} \\
        \cmidrule(lr){3-5} \cmidrule(lr){6-6} \cmidrule(lr){7-9} \cmidrule(l){10-10}
        & & PSNR $\uparrow$ & SSIM $\uparrow$ & LPIPS $\downarrow$ & Total Rb $\downarrow$
        & MANIQA $\uparrow$ & CLIP-IQA $\uparrow$ & MUSIQ $\uparrow$ & Total Rb $\downarrow$ \\
        \midrule
        0 & 0   & 20.59 & 0.6579 & 0.3287 & 2.66 & 0.2770 & 0.4603 & 56.69 & 2.34 \\
        1 & 25  & 22.21 & 0.7346 & 0.3430 & 1.71 & 0.3459 & 0.5091 & 61.26 & 1.63 \\
        2 & 50  & \textbf{23.38} & 0.7643 & 0.3117 & \textbf{1.43} & 0.3735 & 0.5296 & 61.96 & \textbf{1.15} \\
        3 & 75  & 23.03 & 0.7539 & 0.3205 & 1.53 & \textbf{0.3749} & 0.5180 & \textbf{62.44} & 1.49 \\
        4 & 100 & 23.04 & \textbf{0.7647} & \textbf{0.3029} & 1.80 & 0.3714 & \textbf{0.5386} & 61.79 & 1.50 \\
        \bottomrule
    \end{tabular}   
    }
\end{table}

\subsection{Ablation Studies on Rollback Mechanism and Preference Setting}

\textbf{Evaluation on rollback mechanism.} The left of Table \ref{tab:appendix_b1_ablation_rollback_dimension} illustrates that, in nearly all instances, the metrics with the rollback process are superior. Notably, the rollback process differs from \citep{zhu2025intelligent, NEURIPS2025_f0075fe4}, as our method is structured around the experience pool.

\textbf{Evaluation on preference setting.} The right of Table \ref{tab:appendix_b1_ablation_rollback_dimension} clearly demonstrates that fidelity consistently achieves superior FR metrics, while \textit{Perception} dominates in NR metrics across three groups in the MiO100 dataset. It is noteworthy that even when prioritizing perception preference, its FR metrics demonstrate superiority over those of existing state-of-the-art methods.

\begin{table}[htb]
\centering
\caption{Ablation study on rollback mechanism and preference design with \textbf{EvoIR-Agent}.}
\label{tab:appendix_b1_ablation_rollback_dimension}
\resizebox{\textwidth}{!}{%
\begin{tabular}{c ccccccc ccccccc}
\toprule
\multirow{2}{*}{\textbf{Degrad.}} & \multicolumn{7}{c}{\textbf{Rollback Mechanism}} & \multicolumn{7}{c}{\textbf{Preference Setting}} \\
\cmidrule(lr){2-8} \cmidrule(lr){9-15}
& Method & PSNR $\uparrow$ & SSIM $\uparrow$ & LPIPS $\downarrow$ & MANIQA $\uparrow$ & CLIP-IQA $\uparrow$ & MUSIQ $\uparrow$ & Dim & PSNR $\uparrow$ & SSIM $\uparrow$ & LPIPS $\downarrow$ & MANIQA $\uparrow$ & CLIP-IQA $\uparrow$ & MUSIQ $\uparrow$ \\
\midrule
\multirow{2}{*}{Group A}
& w/o Rb. & 23.38 & 0.7410 & 0.3177 & \textbf{0.3660} & \textbf{0.5214} & 61.59 & F & \textbf{23.86} & \textbf{0.7488} & 0.3085 & 0.2833 & 0.4190 & 54.37 \\
& w Rb.   & \textbf{23.86} & \textbf{0.7488} & \textbf{0.3085} & 0.3623 & 0.5204 & \textbf{61.82} & P & 22.31 & 0.7103 & \textbf{0.2831} & \textbf{0.3623} & \textbf{0.5204} & \textbf{61.82} \\
\midrule
\multirow{2}{*}{Group B}
& w/o Rb. & 22.93 & 0.7582 & 0.3162 & \textbf{0.3814} & \textbf{0.5436} & 61.62 & F & \textbf{23.38} & \textbf{0.7643} & 0.3117 & 0.2727 & 0.4247 & 53.28 \\
& w Rb.   & \textbf{23.38} & \textbf{0.7643} & \textbf{0.3117} & 0.3735 & 0.5296 & \textbf{61.96} & P & 22.52 & 0.7392 & \textbf{0.2694} & \textbf{0.3735} & \textbf{0.5296} & \textbf{61.96} \\
\midrule
\multirow{2}{*}{Group C}
& w/o Rb. & 21.98 & 0.6576 & 0.4682 & 0.3278 & 0.4930 & 52.96 & F & \textbf{22.12} & \textbf{0.6588} & 0.4528 & 0.2032 & 0.3201 & 41.36  \\
& w Rb.   & \textbf{22.12} & \textbf{0.6588} & \textbf{0.4528} & \textbf{0.3321} & \textbf{0.4977} & \textbf{55.21} & P & 20.69 & 0.6004 & \textbf{0.4072} & \textbf{0.3321} & \textbf{0.4977} & \textbf{55.21} \\
\bottomrule
\end{tabular}
}
\end{table}

\subsection{Scalability Analysis on Complex Scenarios}
To investigate the scalability in extreme degradation scenarios, we design an exploratory study on two challenging 4-degradation combinations (\textit{Defocus + Haze + JPEG + Rain} and \textit{Dark + Haze + JPEG + Motion Blur}). Specifically, during the experience evolution phase, we introduce a stochastic trajectory dropping strategy with drop rates ranging in $\{0.00, 0.25, 0.50, 0.75\}$ on ~\eqref{search}.

As summarized in Table~\ref{tab:four_deg_drop_rates}, a noticeable performance decay surfaces when the drop rate reaches $0.75$, where the system explores merely $6$ out of $24$ possible permutation trajectories to update the priority function $p$. This demonstrates that our priority modeling exhibits strong robustness against sparse exploration trajectories in more complex cases.

\textbf{Insight on Cascading Error Accumulation.} Crucially, we observe that the overall quantitative scores across all drop rates and one baseline method remain relatively modest. This reveals an intrinsic challenge in long-horizon IRA tasks: in highly entangled degradation chains, the primary bottleneck is \textit{cascading error accumulation} caused by chained sequential tool invocations. Consequently, improving search efficiency alone during the experience evolution stage is insufficient to address extreme combinations; developing error-mitigating inference mechanisms is indispensable for long-horizon agentic restoration, which we envision as our future research focus.

\begin{table}[H]
\centering
\caption{Performance across four degradation combinations under varying trajectory drop rates during experience evolution.}
\label{tab:four_deg_drop_rates}
\small
\setlength{\tabcolsep}{4.2pt}
\begin{tabular}{lccrrrrrr}
\toprule
\multirow{2}{*}{\textbf{Method}} & \multirow{2}{*}{\shortstack{\textbf{Drop}\\\textbf{Rate}}} & \multirow{2}{*}{\shortstack{\textbf{Sampled}\\\textbf{Traj.}}} & \multicolumn{3}{c}{\textbf{Fidelity}} & \multicolumn{3}{c}{\textbf{Perception}} \\
\cmidrule(lr){4-6} \cmidrule(lr){7-9}
& & & PSNR$\uparrow$ & SSIM$\uparrow$ & LPIPS$\downarrow$ & MANIQA$\uparrow$ & CLIP-IQA$\uparrow$ & MUSIQ$\uparrow$ \\
\midrule
AgenticIR & -- & -- & 15.86 & 0.5055 & 0.5575 & 0.1700 & 0.2635 & 38.62 \\
\midrule
\multirow{4}{*}{EvoIR-Agent} 
& 0.00 & 24 / 24 & 17.58 & 0.5379 & 0.5891 & 0.1739 & 0.2586 & 35.03 \\
& 0.25 & 18 / 24 & 17.56 & 0.5397 & 0.5863 & 0.1743 & 0.2621 & 35.09 \\
& 0.50 & 12 / 24 & 17.57 & 0.5405 & 0.5858 & 0.1739 & 0.2598 & 35.21 \\
& 0.75 & 6 / 24  & 17.40 & 0.5326 & 0.5934 & 0.1779 & 0.2623 & 35.62 \\
\bottomrule
\end{tabular}
\end{table}

\subsection{More Qualitative Results}
\begin{figure}[H]
  \centering
  \includegraphics[width=\textwidth]{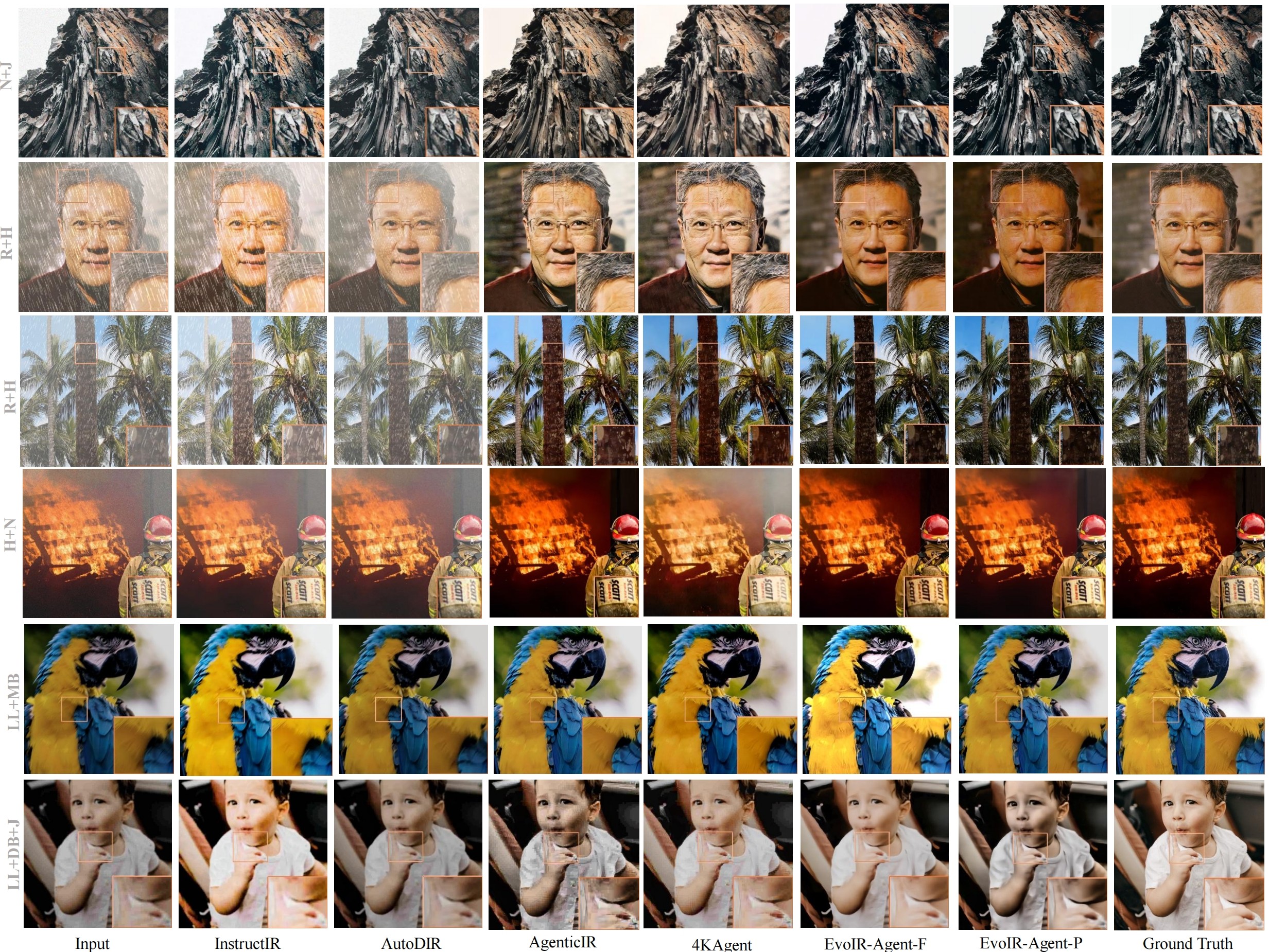}
  \caption{
  More visual comparison examples on MiO100 dataset. \textbf{EvoIR-Agent-F} denoted the fidelity \textit{Preference} requirement, while \textbf{EvoIR-Agent-P} denoted the perception \textit{Preference} requirement.
  }
  \label{fig:qualitative_comparision_appendix}
\end{figure}

\section{Implementation Details}
\label{appendix:experiment_detail}

\subsection{Datasets Detail}
\label{appendix:dataset_detail}
\textbf{Synthetic Dataset:} We adhere to the experimental protocol outlined by AgenticIR \citep{zhu2025intelligent}. Our methodology is data-driven, and to ensure the isolation of the dataset for test-time adaptation, we require data for the purpose of experience evolution.
\begin{itemize}
    \item Experience Evolution Set: To construct our experience pool, we utilize a separate, disjoint set of 3,450 high-quality images provided by MiOIR \citep{kong2024towards}. We leverage the degradation generation pipeline from AgenticIR to generate degraded-gt image pairs for the evolution process.
    \item Evaluation Set: All our evaluations are conducted purely on the MiO100 dataset, which contains totally 1,440 images with 16 degradation coupling scenarios. Group A contains 8 groups with 2 degradation, Group B contains 4 groups with 2 degradation and Group C contains 4 groups with 3 degradation.
\end{itemize}

\textbf{Real-world Dataset:} We are interested in degradation coupling scenarios that are well-suited to the main contributions of the IRAs. To this end, we utilize three groups from a real-world dataset, FoundIR \citep{li2025foundir} for evaluation: motion blur+JPEG, rain+haze, and dark+noise. Each group contains 50 degraded-gt pairs. Unless otherwise specified, this dataset is used for zero-shot generalization evaluation.

\subsection{Tool Set}
\label{appendix:tool_set}
We leverage the methods in \citep{zhu2025intelligent} during the \textit{Perception} and \textit{Reflection} processes. The entire experiment is conducted with two NVIDIA A6000 GPUs. Since the present framework is flexible to different tools, this property is examined in two tool sets. All the ablation studies are conducted on setting II without specification.

In the \textbf{EvoIR-Agent-I}, the configuration of the tool settings in previous agents \citep{zhu2025intelligent, jiang2025multi, lu2025simplecall} is adhered to.
\begin{itemize}
    \item \textbf{Brightening:} CLAHE~\citep{reza2004realization}, Gamma correction ($\gamma = 2/3$), constant shift.
    
    \item \textbf{Defocus Deblurring:} DRBNet~\citep{ruan2022learning}, IFAN~\citep{lee2021iterative}, Restormer~\citep{zamir2022restormer}.
    
    \item \textbf{JPEG Artifact Removal:} SwinIR~\citep{liang2021swinir} (quality factor 40), FBCNN~\citep{jiang2021towards} (quality factors 90 and 5, and blind setting).
    
    \item \textbf{Denoising:} SwinIR~\citep{liang2021swinir}, MAXIM~\citep{tu2022maxim}, MPRNet~\citep{zamir2021multi}, Restormer~\citep{zamir2022restormer}, X-Restormer~\citep{chen2024comparative}.
    
    \item \textbf{Deraining:} MAXIM~\citep{tu2022maxim}, MPRNet~\citep{zamir2021multi}, Restormer~\citep{zamir2022restormer}, X-Restormer~\citep{chen2024comparative}.
    
    \item \textbf{Motion Deblurring:} MAXIM~\citep{tu2022maxim}, MPRNet~\citep{zamir2021multi}, Restormer~\citep{zamir2022restormer}, X-Restormer~\citep{chen2024comparative}.
    
    \item \textbf{Dehazing:} MAXIM~\citep{tu2022maxim}, X-Restormer~\citep{chen2024comparative}, RIDCP~\citep{wu2023ridcp}, DehazeFormer~\citep{song2023vision}.

    \item \textbf{Super Resolution:} DiffBIR \citep{lin2024diffbir}, HAT \citep{chen2025hat}, SwinIR \citep{liang2021swinir}, X-Restormer~\citep{chen2024comparative}.
    
\end{itemize}

In the \textbf{EvoIR-Agent-II}, the tool settings from prior agents \citep{NEURIPS2025_f0075fe4, zhang2026tir} are utilized.

\begin{itemize}
    \item \textbf{Brightening:} CLAHE~\citep{reza2004realization}, Constant Shift (C=40), DiffPlugin \citep{liu2024diff}, FourierDiff \citep{lv2024fourier}, Gamma Correction ($\gamma = 2/3$), MAXIM \citep{tu2022maxim}.
    
    \item \textbf{Defocus Deblurring:} ConvIR \citep{cui2024revitalizing}, DiffPlugin \citep{liu2024diff}, DRBNet~\citep{ruan2022learning}, IFAN~\citep{lee2021iterative}, LaKDNet \citep{cui2024omni}, Restormer~\citep{zamir2022restormer}.

    \item \textbf{JPEG Artifact Removal:} FBCNN~\citep{jiang2021towards} (QF=5), FBCNN~\citep{jiang2021towards} (QF=90), FBCNN~\citep{jiang2021towards} (BQF), SwinIR~\citep{liang2021swinir} (QF=40).
    
    \item \textbf{Denoising:} MAXIM~\citep{tu2022maxim}, MPRNet~\citep{zamir2021multi}, NAFNet \citep{chen2022simple}, Restormer~\citep{zamir2022restormer}, X-Restormer~\citep{chen2024comparative}, SwinIR~\citep{liang2021swinir}.

    \item \textbf{Deraining:} DiffPlugin \citep{liu2024diff}, MAXIM~\citep{tu2022maxim}, MPRNet~\citep{zamir2021multi}, Restormer~\citep{zamir2022restormer}, X-Restormer~\citep{chen2024comparative}.
    
    \item \textbf{Dehazing:} DehazeFormer~\citep{song2023vision}, DiffPlugin \citep{liu2024diff}, MAXIM~\citep{tu2022maxim}, RIDCP~\citep{wu2023ridcp}, X-Restormer~\citep{chen2024comparative}.
    
    \item \textbf{Motion Deblurring:} EVSSM \citep{kong2025efficient}, LaKDNet \citep{cui2024omni}, MAXIM~\citep{tu2022maxim}, MPRNet~\citep{zamir2021multi}, NAFNet \citep{chen2022simple}, Restormer~\citep{zamir2022restormer}, X-Restormer~\citep{chen2024comparative}.
    
    \item \textbf{Super Resolution:} DiffBIR \citep{lin2024diffbir}, DRCT \citep{hsu2024drct}, HAT-L \citep{chen2025hat}, HAT-GAN, HMA \citep{chu2024hmanet}, OSEDiff \citep{wu2024one}, PiSA-SR \citep{sun2025pixel}, SwinIR \citep{liang2021swinir}, SwinIR (Real-ISR), SwinFIR \citep{zhang2022swinfir}, X-Restormer~\citep{chen2024comparative}.
    
\end{itemize}

\subsection{Experience Pool Setting}
\textbf{Implementation Details \& Hyperparameters.} 
Our framework is implemented with the following hyperparameter configurations:
\begin{itemize}
    \item \textbf{Retrieval \& Evolution:} The Top-K recall (Sec.~\ref{subsec:experience_driven_inference_workflow}) is set to $3$. For experience evolution (Sec.~\ref{subsec:self-evolving_experience_mechanism}), we set the trigger threshold batch size $B$ to $25$, the significance level $\alpha$ to $0.975$, and the mini-batch size to $12$.
    \item \textbf{IQA Metrics:} The metrics set for \textit{Fidelity} consist of PSNR, SSIM, LPIPS, and DISTS. The \textit{Perception} metrics set, utilized to support the \textit{Preference} in Sec.~\ref{subsec:experience_formulation}, includes MANIQA, MUSIQ, BRISQE, CLIPIQA, NIQE, and NIMA.
\end{itemize}

\textbf{MLLM Instantiation.} 
The estimated API cost per round for the agent-driven pattern profile learning system is reported in Table~\ref{tab:appendix_c2_api_cost}.
\begin{itemize}
    \item \textbf{Base Inference:} Qwen3-VL-Flash~\citep{yang2025qwen3} is adopted for standard, general-purpose multimodal operations.
    \item \textbf{Experience Evolution:} To ensure deep image interpretation and high-quality description generation during fine-grained experience evolution, we employ the more capable Qwen3-Plus~\citep{yang2025qwen3} foundation model. 
\end{itemize}

\begin{table}[htb]
    \centering
    \setlength{\tabcolsep}{12pt} 
    \renewcommand{\arraystretch}{1.2}
    
    \caption{Detailed API usage and estimated cost for formulating fine-grained experience within a single mini-batch in \textbf{EvoIR-Agent}.}
    \label{tab:appendix_c2_api_cost}

    \begin{tabular}{lcccc}
        \toprule
        \multirow{2}{*}{\textbf{Model Name}} & \textbf{Calls /} & \multicolumn{2}{c}{\textbf{Avg. Tokens / Call}} & \textbf{Total Cost} \\
        \cmidrule(lr){3-4}
        & \textbf{Batch} & \textbf{Input} & \textbf{Output} & \textbf{per Batch (\$)} \\
        \midrule
        Qwen3-VL-Flash & $\sim$35 & $\sim$950 & $\sim$150 & 0.0189 \\
        Qwen3-Plus     & $\sim$12 & $\sim$1,300 & $\sim$400 & 0.0931 \\
        \midrule
        \textbf{Total} & \textbf{47} & \multicolumn{2}{c}{--} & \textbf{0.1120} \\
        \bottomrule
    \end{tabular}
\end{table}

\subsection{Prompts}

\begin{rawpromptbox}[\textit{Insight} Evolving Prompt]
Based on the following Bradley-Terry-Davison (BTD) probabilistic model results for multi-degradation image restoration tasks (Preference: {preference}), extract key insights regarding the degradation removal order and recommendations:

{combined_text}

Your conclusions and insights:
\end{rawpromptbox}

\begin{rawpromptbox}[Multi-Agent Debate \textit{Role} Prompt]
Your role is to act as the {role}. The theme is: "Is this a good enough degradation pattern?". This means you are skilled at processing information from both yourself and others to determine the next appropriate action without executing it. You are currently in the thinking stage.

Historical information (your current-stage reasoning & external tool results): {context}.

You have access to external tools represented as Python function prototypes, which you may invoke to support your debate competition:
- generate_groups() # Used to generate image degradation patterns solely from textual information in the `[Start Round]`.
- validate_current_group(traj_id: List[int]) # Used to validate the current unconvinced image degradation pattern (intra-group: within the same degradation pattern) using specified trajectory IDs. (Requirements: 1 unconvinced traj_id + 1~3 convinced traj_ids; maximum of 3 traj_ids in total)
- validate_other_group(traj_id: List[int]) # Used to validate another unconvinced image degradation pattern (inter-group: between two degradation patterns) using specified trajectory IDs. (Requirements: 1 unconvinced traj_id + 1~3 convinced traj_ids; maximum of 3 traj_ids in total)
- finish() # Used when you think you have obtained the final answer.

Your response must strictly follow this format. It is forbidden to include any output beyond these two sections:
Thought: [Your brief reasoning process (under 200 words), including problem analysis and determination of the next action by carefully reviewing the historical information from your opponent and your current-stage reasoning.]
Action: [The Python function call you decide to execute in the next step]
\end{rawpromptbox}

\begin{rawpromptbox}[Multi-Agent Debate \textit{Action} Prompt]
You want to ensure that all degradation patterns you propose are credible and accurate enough to withstand challenges in a debate competition. The theme is: `Is this a good enough degradation pattern?`.

Currently, there is an ongoing process defined as "semantic clustering".
The goal of semantic clustering is to categorize degradation modes according to the following rules:
1. Within each degradation pattern category, the semantic representations of the degradation pattern text must be as similar as possible.
2. Between different degradation pattern categories, the semantic representations of the degradation pattern text must be significantly distinct.
3. Under the premise of satisfying Rules 1 and 2, the tool ranking information of the top three (or top 2, if only two exist) within each category must be as close as possible.
4. Under the premise of satisfying Rules 1 and 2, the tool ranking information of the top three (or top 2, if only two exist) between different categories must be as distant as possible.
5. The number of degradation pattern categories is flexible, but it must avoid redundancy and satisfy Rules 1 to 4.
6. For uncertain trajectory information that cannot be confidently classified considering both semantic representation and ranking requirements, explicitly specify which trajectories are uncertain and require further verification inside `<unconvinced_info></unconvinced_info>` or `<verify_1></verify_1>.

You are provided with the current degradation pattern categories derived solely from textual information: ... 
Some categories contain trajectories with uncertain classifications. Now, you will perform a further validation step using the image information provided as: ...

You can provide two types of judgments:
1. Positive: Confirm that the trajectory fits the degradation pattern category based on the image data.
2. Negative: Determine that the trajectory does not fit the current degradation pattern category; instead, select a more appropriate category based on textual semantic similarity and specify that cross-category verification is needed.

Finally, you must retain all previous outputs inside `<convinced_info></convinced_info>` (as well as existing `<unconvinced_info></unconvinced_info>` and `<verify_n></verify_n>` tags) without modifying or deleting them. For the current verification, append your result inside an XML tag with an incremental sequence number (e.g., `<verify_1></verify_1>`). 

Your output must strictly follow this format:
<convinced_info>Degradation Pattern 1: The pure semantic representation of the degradation pattern text information is... (Aspect 1: Semantic representation of the degradation pattern, without tool names or removal priorities, at least 15 words). Included trajectories are Traj1, Traj3..., reasons are: (Combine three aspects: Semantic representation of the degradation pattern & Top-3 removal priority rankings [Top-2 if only two] & Tool names)...</convinced_info><unconvinced_info>Among these trajectories, the classifications of Traj1, ... are not fully convincing because..., requiring further verification.</unconvinced_info><verify_1>After combined text and image validation, Traj3 indeed belongs to Degradation Pattern 1, because...</verify_1>
\end{rawpromptbox}

\begin{rawpromptbox}[Pattern Profile Operation \textit{Function} Prompt]
You are an expert in planning and knowledge management, tasked with updating degradation patterns stored within a vector database. Your objective is to analyze a given set of "new degradation patterns" alongside the "existing degradation patterns" currently in the database, and determine the precise atomic operations required to keep the database accurate and up-to-date.

Specifically, for each new pattern, decide which of the following actions applies:
- add: Introduce a completely new pattern to the database because it does not overlap with existing ones. Use this operation to balance accuracy and redundancy.
- merge: Apply when an existing pattern fully subsumes a new pattern. Retain the existing pattern but enrich it with additional information from the new one.
- replace: Apply when a new pattern subsumes an existing pattern, meaning the new pattern's semantics are broader or more accurate, and thus should replace the existing pattern.

You will receive a task description that specifies the particular patterns to analyze. If you receive any feedback on your output, revise your operations accordingly while maintaining the same output format.

Current task description:
Degradation type: {degradation_type}
New degradation patterns: {new_dp}

Existing degradation patterns in DB:
<none>

Previous plans:
<none>

Previous feedback:
<none>

Your output must be formatted as a Python list of strings, where each string represents a single atomic operation. It is strictly forbidden to perform multiple operations on the same existing degradation pattern. If an operation on a pattern is unsatisfactory, choose an alternative degradation pattern.

The format for each element is:
"<new pattern digit only> | <action> | <existing pattern digit only (optional)>"

- Include the existing pattern digit only when relevant (for merge or replace actions).
- For add actions, the existing pattern field must be omitted.


Example output:
["1 | merge | 2", "2 | add"]
\end{rawpromptbox}

\end{document}